\documentclass{article}
\usepackage{natbib}
\usepackage{amsmath,amssymb,amsthm,mathtools,bm}
\usepackage{hyperref}
\usepackage{url}
\usepackage{amsmath,amssymb,amsfonts}
\usepackage{tikz}
\usepackage{pgfplots}
\usepackage{cleveref}
\usetikzlibrary{pgfplots.groupplots}
\pgfplotsset{compat=1.18}
\usepackage{algorithm}
\usepackage{algpseudocode}
\usepackage{caption}
\usepackage{booktabs}
\usepackage{multirow}
\usepackage{tabularx}
\usepackage{graphicx}
\usepackage{subcaption}
\usepackage[letterpaper, margin=1in]{geometry}

\title{Directional-Clamp PPO}
\author{}       
\date{}         
\begin{document}
\newtheorem{lemma}{Lemma}

\maketitle
\author{
\begin{center}
\begin{tabular}{c@{\hspace{3cm}}c}
\begin{tabular}{c}
\textbf{Gilad Karpel} \\[-2pt]
\small Technion – Israel Institute of Technology \\[4pt]
\small \texttt{gilad.karpel@campus.technion.ac.il} \\[8pt]

\textbf{Shoham Sabach} \\[-2pt]
\small Technion, Amazon AGI \\[4pt]
\small \texttt{ssabach@technion.ac.il}
\end{tabular}
&
\begin{tabular}{c}
\textbf{Ruida Zhou} \\[-2pt]
\small Amazon AGI \\[4pt]
\small \texttt{zruida@amazon.com} \\[8pt]

\textbf{Mohammad Ghavamzadeh} \\[-2pt]
\small Amazon AGI \\[4pt]
\small \texttt{ghavamza@amazon.com}
\end{tabular}
\end{tabular}
\end{center}
}

\begin{center}
\begin{abstract}
    Proximal Policy Optimization (PPO) is widely regarded as one of the most successful deep reinforcement learning algorithms, known for its robustness and effectiveness across a range of problems. 
    The PPO objective encourages the importance ratio between the current and behavior policies to move to the ``right'' direction -- starting from importance sampling ratios equal to 1, increasing the ratios for actions with positive advantages and decreasing those with negative advantages. A clipping function is introduced to prevent over-optimization when updating the importance ratio in these ``right'' direction regions. Many PPO variants have been proposed to extend its success, most of which modify the objective’s behavior by altering the clipping in the ``right'' direction regions. However, due to randomness in the rollouts and stochasticity of the policy optimization, we observe that the ratios frequently move to the ``wrong'' direction during the PPO optimization. This is a key factor hindering the improvement of PPO, but it has been largely overlooked. To address this, we propose the Directional-Clamp PPO algorithm (DClamp-PPO), which further penalizes the actions going to the strict ``wrong'' direction regions, where the advantage is positive (negative) and importance ratio falls below (above) $1 - \beta$ ($1+\beta$),  
    for a tunable parameter $\beta \in (0, 1)$. The penalty is by enforcing a steeper loss slope, i.e., a clamp, in those regions. We demonstrate that DClamp-PPO consistently outperforms PPO, as well as its variants, by focusing on modifying the objective's behavior in the ``right'' direction, across various MuJoCo environments, using different random seeds. The proposed method is shown, both theoretically and empirically, to better avoid ``wrong'' direction updates while keeping the importance ratio closer to 1. 
\end{abstract}
\end{center}

\section{Introduction}
Deep Reinforcement Learning (DRL) algorithms have achieved remarkable success in a variety of complex domains, transforming fields such as robotics~\citep{kober2013reinforcement}, gaming~\citep{mnih2015human,silver2016mastering}, and autonomous systems~\citep{gu2017deep}. These breakthroughs demonstrate the ability of models to learn effective strategies through trial and error in environments with uncertain dynamics and partial observability. DRL has also been employed in autonomous driving applications~\citep{sallab2017deep}, highlighting its potential to solve a wide range of real-world challenges. A very popular class of DRL algorithms, which is called policy gradient methods, focuses on directly optimizing a parameterized policy to maximize its expected cumulative reward. The main advantage of policy gradient methods is their directness and ability to handle high-dimensional and continuous action spaces. Typically, policy gradient methods use an estimator of the policy gradient and plug it into stochastic gradient ascent. The convergence properties of the policy gradient methods are strongly influenced by the choice of step size: small values imply a slow convergence rate, while large values may lead to oscillation or even divergence of the policy parameters~\citep{pirotta2013adaptive}. The well-known \textit{Trust-Region Policy Optimization} algorithm (TRPO)~\citep{schulman2015trust} addresses the issue with sensitivity to step sizes. TRPO theoretically guarantees monotonic policy improvement by constraining the Kullback–Leibler (KL) divergence between the old policy and the new policy. However, enforcing this constraint is computationally expensive, as it requires a second-order approximation of the KL divergence and the use of a conjugate gradient method, making TRPO difficult to scale. 
\medskip

To overcome these limitations, \textit{Proximal Policy Optimization} (PPO)~\citep{schulman2017proximal} introduces a simpler clipping mechanism that approximates TRPO’s trust-region constraint. Throughout this paper, we refer to the PPO objective function as the PPO-clip objective function. The key idea is to constrain the importance ratio, i.e., the current policy divided by the trajectory sampling old policy (rollout policy), within a small interval $[1-\epsilon, 1+\epsilon]$ to avoid over-optimization. Specifically, the target is to increase (decrease) the ratio of the actions with positive (negative) advantage, 
which defines the ``right'' direction to optimize. The PPO objective function is clipped with zero gradient if the ratios are over-optimized too much along the ``right'' direction, i.e., ratios greater than $1+\epsilon$ for positive advantage or smaller than $1-\epsilon$ for negative advantage. By this simple clipping, the PPO objective creates a natural trust-region. Due to its simplicity and efficiency, PPO is an iconic policy gradient algorithm that has been applied in many different fields and shown great stability accross domains, including games~\citep{berner2019dota},  healthcare~\citep{yu2021reinforcement}, finance~\citep{ye2020reinforcement}, UAV control~\citep{koch2019reinforcement}, and more recently alignment of large language models (LLM)~\citep{ouyang2022training}. The PPO objective is designed to update the ratios in the right direction; however, we find that a substantial portion of the importance sampling ratios still drift in the wrong direction -- lowering the probability of advantageous actions and raising that of disadvantageous ones -- during the PPO training, as we will demonstrate in Figure \ref{fig:ratio_histograms}. This ``wrong-directional'' drift stems from the stochastic nature of the training process: the trajectories are randomly sampled, the advantage estimates can have high variance, and the policy updates are performed on small mini-batches.
\medskip

To address this issue, we introduce Directional-Clamp PPO (DClamp-PPO), which augments the standard PPO objective with a targeted penalty that more aggressively suppresses wrong-direction updates. The \emph{directional} component applies this penalty selectively, activating it only when the ratio stays in the strictly wrong-direction region, i.e., ratio less than $(1-\beta)$ for positive advantage or ratio greater than $(1+\beta)$ for negative advantage, where the $\beta$ is a parameter to control this region. As illustrated in Figure \ref{fig:method}, the \emph{clamp} component sharpens the loss landscape in these regions, effectively pulling the importance ratio back toward~1 and aligning policy updates more closely with the trust-region principle. Through this mechanism, DClamp-PPO substantially reduces harmful deviations, yielding both theoretical stability guarantees and empirical improvements over PPO and its strongest variants across diverse MuJoCo benchmarks.
\medskip

\noindent
Our contributions are summarized as follows:
    \begin{enumerate}
    \item We observe that a proportion of importance sampling ratios drift into the wrong direction during PPO optimization, which can degrade its performance but has been largely overlooked. 
    \item We propose DClamp-PPO, a simple, intuitive, and effective modification on the objective function of PPO without any overhead to the standard PPO pipeline.
    \item Theoretically, we show that under some mild conditions, the DClamp-PPO update keeps the ratio closer to 1, thereby reducing harmful deviations and aligning updates within the trust-region.
    \item DClamp-PPO consistently outperforms PPO and other competitive baselines reported to achieve state-of-the-art performance for various MuJoCo environments. This could motivate future research on the ``wrong'' direction region of the policy gradient algorithms.
\end{enumerate}

\section{Background}

We consider a discounted MDP $\mathcal{M}=(\mathcal{S},\mathcal{A},P,r,\rho_0,\gamma)$ with state space $\mathcal{S}$, action space $\mathcal{A}$, transition probability $P:\mathcal{S}\times\mathcal{A}\times\mathcal{S}\to[0,1]$, reward function $r:\mathcal{S}\times\mathcal{A}\to\mathbb{R}$, discount factor $\gamma\in(0,1)$, and initial state distribution $\rho_0$ over $\mathcal{S}$. A stochastic Markov policy $\pi(\cdot|s) \in \Delta_{\mathcal{A}}$ induces trajectories $(s_0,a_0,r_0,s_1,\ldots)$ where $s_0\sim\rho_0$, $a_t\sim\pi(\cdot|s_t)$, and $s_{t+1}\sim P(\cdot|s_t,a_t)$. The goal is to maximize the expected discounted return $ J(\pi) := \mathbb{E}_{s_0 \sim \rho, \pi,P}\left[\sum_{t=0}^{\infty}\gamma^t\,r(s_t,a_t)\right]$. The state-value and action-value functions under $\pi$ are defined as 
$$V^\pi(s) :=\mathbb{E}_{\pi,P}\!\left[\sum_{t=0}^{\infty}\gamma^t r(s_t,a_t)\,\bigg|\,s_0=s\right],\qquad Q^\pi(s,a) :=\mathbb{E}_{\pi,P}\!\left[\sum_{t=0}^{\infty}\gamma^t r(s_t,a_t)\,\bigg|\,s_0=s,a_0=a\right],$$
and the advantage function is defined by $A^\pi(s,a):=Q^\pi(s,a)-V^\pi(s)$. In practice, policies are drawn from a parameterized family  $\{\pi_\theta : \theta \in \Theta\}$, where $\pi_\theta(a|s)$ denotes the probability
of action $a$ in state $s$ under parameters $\theta$.
 
\paragraph{Policy Gradient} The MDP objective $J(\pi_{\theta})$ is differentiable with respect to the parameterized policy, and its gradient is obtained by the policy gradient theorem \citep{sutton1999policy} as
\begin{align*}
    (1-\gamma)\nabla_\theta  J(\pi_\theta) & = 
\mathbb{E}_{s \sim \rho_{\pi_\theta}, a \sim \pi_{\theta}(\cdot | s)} \!\left[
  \nabla_\theta \log \pi_\theta(a|s)\, A^{\pi_\theta }(s, a)
\right]
\end{align*}
where $\rho_{\pi}(s):= (1-\gamma)\sum_{t \geq 0} \gamma^t \mathbb{P}(s_t = s)$ is the discounted state visiting distribution under any policy $\pi$, and $\pi_{\text{old}}$ is a rollout policy that can potentially be different from $\pi_{\theta}$. The expectation can be estimated via rollout trajectories, and the advantage function can be estimated by various estimators in the pursuit of less variance and bias. A well-adopted choice for the advantage estimator is Generalized Advantage Estimator (GAE), which estimate $A^{\pi}(s_t, a_t)$ by 
\begin{align}
\hat{A}(s_t,a_t) := \sum_{l=0}^{T-t-1} (\gamma \lambda)^l \, \delta_{t+l}, 
\qquad \text{where } \;\; \delta_t = r_t + \gamma V_{\phi}(s_{t+1}) - V_{\phi}(s_t),
\label{eq:GAE}
\end{align}
where $V_{\phi}$ is a critic function that approximates the value function \(V^{\pi}(s)\). 

\paragraph{Trust-Region Policy Optimization} TRPO~\cite{schulman2015trust} was proposed to reuse the rollout data and increase the stability of policy gradient methods. It tackles the policy optimization problem via iteratively solving the following constraint surrogate:
\begin{align*}
\max_{\theta} \;\; 
\hat{\mathbb{E}}_{s,a}\!\left[
  \tfrac{\pi_\theta(a|s)}{\pi_{\theta_{\text{old}}}(a|s)} \,\hat{A}(s,a)
\right], \qquad \text{s.t.} \quad 
\hat{\mathbb{E}}_{s,a}\!\left[
  D_{\mathrm{KL}}\!\big(\pi_{\theta_{\text{old}}}(\cdot|s)\,\|\,\pi_\theta(\cdot|s)\big)
\right] \;\leq\; \delta,
\end{align*}
where $D_{\mathrm{KL}}$ is the KL divergence and $\hat{\mathbb{E}}_{s,a}[\cdot]$ denotes the empirical expectation over a batch of sampled state–action pairs. This constrained optimization surrogate theoretically guarantees monotonic policy improvement given a sufficiently large amount of rollout samples and a small trust-region $\delta$. However, practically solving this optimization is computationally expensive, as it requires a second-order KL approximation with the Fisher information matrix and conjugate gradient descent for the quadratic program. 

\paragraph{Proximal policy gradient and its variants} To overcome this challenge, \cite{schulman2017proximal} proposed PPO as an alternative that relies only on first-order optimization. There are different variations of the PPO formulation, but we take the most adopted one in practice, the PPO(-clip), in which the constrained optimization of TRPO is replaced by a simple clipped surrogate function to maximize 
\begin{align}
    \mathcal{J}_\text{PPO}(\theta) = \hat{\mathbb{E}}_{s,a}\Big[
  \min\big(
    w_{\theta}(s,a)\,\hat{A}(s,a),\;
    \operatorname{clip}(w_{\theta}(s,a),\,1-\epsilon,\,1+\epsilon)\,\hat{A}(s,a)
  \big)
\Big], \label{eqn:PPO}
\end{align}
where $w_\theta(s,a)=\frac{\pi_\theta(a_t|s_t)}{\pi_{\theta_{\text{old}}}(a_t|s_t)}$ and \(\epsilon>0\) is the clipping hyperparameter. 
\medskip

There are many PPO variants that introduce regularization terms other than KL or adaptively update the trust-region. The PPO-based variants mostly focus on the clipping part of the surrogate loss, since it is believed to be the essence of the PPO's success. Here, we introduce two PPO variants that mathematically contradict each other due to different perspectives on the clip part of PPO. 
\medskip

Leaky PPO~\citep{han2024leaky} believes the clipping may slow down the training due to the vanishing gradient information outside the clipping band, and thus, proposes the following surrogate function:
\begin{align}
\mathcal{J}_{\text{Leaky-PPO}}(\theta)
= \hat{\mathbb{E}}_{s,a}\!\left[
\min\!\Big(
w_{\theta}(s,a)\,\hat{A}(s,a),\;
f_{\text{Leaky}}\big(w_{\theta}(s,a),\epsilon,\alpha\big)
\,\hat{A}(s,a)
\Big)
\right],
\label{eqn:leaky-ppo}
\end{align}
where $\alpha \in [0,1)$ is a hyperparameter controlling the slope of the clip that leaks the gradient information, and 
\begin{equation*}
f_{\text{Leaky}}\big(w_{\theta}(s,a),\epsilon, \alpha\big) =
\begin{cases}
\alpha\,w_{\theta}(s,a) + (1-\alpha)(1-\epsilon) & \text{if } \; w_{\theta}(s,a) \leq 1-\epsilon, \\
\alpha\,w_{\theta}(s,a) + (1-\alpha)(1+\epsilon) & \text{if } \; w_{\theta}(s,a) \geq 1+\epsilon, \\
w_{\theta}(s,a) & \text{otherwise}.
\end{cases}
\end{equation*}

On the other hand, \cite{wang2020truly} aimed at further increasing PPO's stability via enforcing stricter trust-region constraint using the following surrogate function:
\begin{align}
    \mathcal{J}_{\text{PPO-RB}}(\theta) 
= \hat {\mathbb{E}}_{s,a}\!\left[
  \min\!\Big(
    w_{\theta}(s,a)\hat{A}(s,a), \;
    f_{\text{RB}}\big(w_{\theta}(s,a), \epsilon, \alpha\big)\,\hat{A}(s,a)
  \Big)
\right], \label{eqn:ppo-rb}
\end{align}
with $\alpha > 0$ and
\[
f_{\text{RB}}\big(w_{\theta}(s,a), \epsilon, \alpha\big) =
\begin{cases}
-\alpha\,w_{\theta}(s,a) + (1+\alpha)(1-\epsilon) & \text{if } \; w_{\theta}(s,a) \leq 1-\epsilon, \\
-\alpha\,w_{\theta}(s,a) + (1+\alpha)(1+\epsilon) & \text{if } \; w_{\theta}(s,a) \geq 1+\epsilon, \\
w_{\theta}(s,a) & \text{otherwise}.
\end{cases}
\]

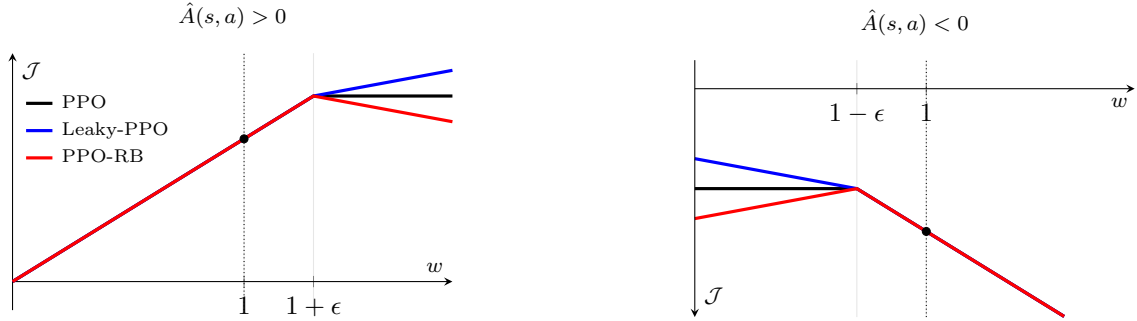
\begin{figure}[t]
\centering

\def\eps{0.30}      
\def\alph{0.30}    
\pgfplotsset{compat=1.18}
\usetikzlibrary{decorations.pathreplacing}

\begin{minipage}{0.45\textwidth}
\centering
\begin{tikzpicture}
\pgfmathsetmacro{\A}{1}             
\pgfmathsetmacro{\Xm}{1-\eps}
\pgfmathsetmacro{\Xp}{1+\eps}
\pgfplotsset{
  legend image code/.code={
    \draw[#1, line width=1.2pt] (0cm,0cm) -- (0.35cm,0cm);
  }
}

\begin{axis}[
  xmin=0, xmax=1.9, ymin=-0.2, ymax=1.6,
  axis x line=middle,
  axis y line=middle,
  axis line style={-stealth, line width=0.4pt},
  width=\textwidth, height=5cm,
  clip=true,
  grid=both, major grid style={gray!20},
  xlabel={\(w\)}, xlabel style={font=\footnotesize},
  ylabel={\(\mathcal J\)}, ylabel style={font=\footnotesize},
  title={\(\hat{A}(s,a)>0\)}, title style={font=\footnotesize},
  xtick={0,1,1+\eps}, xticklabels={$0$,$1$,$1+\epsilon$},
  ytick=\empty,
  legend style={
    font=\scriptsize,
    at={(0.02,0.87)},      
    anchor=north west,
    draw=none,
    fill=none,
    column sep=1pt,
    legend cell align=left,
    inner sep=1pt
  }
]

\addplot[densely dotted, forget plot] coordinates {(1,-0.2) (1,1.6)};

\addplot[only marks, mark=*, mark size=1.5pt, black, forget plot]
  coordinates {(1,\A)};

\addplot[black, very thick, domain=0:1.9, samples=400]
{ min( \A*x, \A*( x < (1-\eps) ? (1-\eps) : ( x > (1+\eps) ? (1+\eps) : x ) ) ) };

\addplot[very thick, domain=0:1.9, samples=400, mark=none, color=blue]
{ min( \A*x,
       \A*( x < (\alph*x + (1-\alph)*(1-\eps)) ? (\alph*x + (1-\alph)*(1-\eps))
            : ( x > (\alph*x + (1-\alph)*(1+\eps)) ? (\alph*x + (1-\alph)*(1+\eps)) : x ) )
) };

\addplot[very thick, domain=0:1.9, samples=400, mark=none, color=red]
{ min( \A*x,
       \A*( x <= (1-\eps) ? ( -\alph*x + (1+\alph)*(1-\eps) )
          : ( x >= (1+\eps) ? ( -\alph*x + (1+\alph)*(1+\eps) ) : x ) )
     )
};

\addlegendimage{black, very thick, mark=none}
\addlegendentry{PPO}

\addlegendimage{orange, very thick, mark=none}
\addlegendentry{Leaky-PPO}

\addlegendimage{red, thick, dashed, mark=none}
\addlegendentry{PPO-RB}

\end{axis}
\end{tikzpicture}
\end{minipage}
\hfill
\begin{minipage}{0.45\textwidth}
\centering
\begin{tikzpicture}
\pgfmathsetmacro{\A}{-1}             
\pgfmathsetmacro{\Xm}{1-\eps}
\pgfmathsetmacro{\Xp}{1+\eps}

\begin{axis}[
  xmin=0, xmax=1.9, ymin=-1.6, ymax=0.2,
  axis x line=middle,
  axis y line=middle,
  axis line style={draw=none},  
  tick style={line width=0.4pt},
  clip=true,                     
  every axis plot/.append style={
    restrict x to domain=0:1.9,
    restrict y to domain=-1.6:0.2
  },
  width=\textwidth, height=5cm,
  grid=both, major grid style={gray!20},
  title={\(\hat{A}(s,a)<0\)}, title style={font=\footnotesize},
  xtick={0,1-\eps,1},
  xticklabels={$0$,$1-\epsilon$,$1$},
  ytick=\empty,
  legend style={font=\scriptsize, at={(0.1,0.98)}, anchor=north west, draw=none, fill=none},
  after end axis/.code={
    \draw[-stealth, line width=0.4pt] (axis cs:0,0) -- (axis cs:1.9,0);  
    \draw[-stealth, line width=0.4pt] (axis cs:0,0.2) -- (axis cs:0,-1.6);
    \node[font=\footnotesize, anchor=north] at (axis cs:1.835,0) {\(w\)};
    \node[font=\footnotesize, anchor=west] at (axis cs:0,-1.45)   {\(\mathcal{J}\)};
  },
]

\addplot[densely dotted] coordinates {(1,-1.6) (1,0.2)};
\addplot[only marks, mark=*, mark size=1.5pt, black] coordinates {(1,\A)};

\addplot[black, very thick, domain=0:1.9, samples=400]
{ min( \A*x, \A*( x < (1-\eps) ? (1-\eps) : ( x > (1+\eps) ? (1+\eps) : x ) ) ) };

\addplot+[very thick, domain=0:1.9, samples=400, mark=none, color=blue]
{ min( \A*x,
       \A*( x < (\alph*x + (1-\alph)*(1-\eps)) ? (\alph*x + (1-\alph)*(1-\eps))
            : ( x > (\alph*x + (1-\alph)*(1+\eps)) ? (\alph*x + (1-\alph)*(1+\eps)) : x ) )
) };

\addplot+[very thick, domain=0:1.9, samples=400, mark=none, color=red]
{ min( \A*x,
       \A*( x <= (1-\eps) ? ( -\alph*x + (1+\alph)*(1-\eps) )
          : ( x >= (1+\eps) ? ( -\alph*x + (1+\alph)*(1+\eps) ) : x ) )
     )
};

\end{axis}
\end{tikzpicture}
\end{minipage}
\caption{The surrogate objective of PPO, Leaky PPO and PPO-RB as a function of the likelihood ratio $w_{\theta}(s,a)$ for positive advantage (left) and negative advantage (right). The black dot denotes $w_{\theta}(s,a) = 1$.}
\label{fig:all-algo}
\end{figure}

Figure~\ref{fig:all-algo} shows the difference between the surrogate objectives of PPO and these two algorithms. From Figure \ref{fig:all-algo}, we notice that both Leaky-PPO and PPO-RB attribute the success and potential improvements of PPO to the clip function, which even lead to their contradicting algorithmic designs. We do not judge which of these variants is better as it depends on the situation, such as the environment and hyper-parameters. While these approaches specifically modify the objective’s behavior by adjusting the clipping mechanism in the “right" direction regions, we highlight that there also exists a ``wrong'' direction that can significantly improve PPO, but has been largely overlooked.

\section{Main Results} \label{sec:main-results}

This section quantifies and addresses a directional failure mode of PPO. Section~\ref{subsec:IR-WD} explicitly defines a ``right'' and ``wrong'' direction updates, illustrates them on the PPO surrogate (Figure~\ref{fig:right-wrong-direction-ppo-loss}), and empirically shows substantial mass in the wrong direction (Figure~\ref{fig:ratio_histograms}). Motivated by these observations, Section~\ref{sec:DClamp-PPO} introduces the notion of a strict wrong direction and proposes Directional-Clamp PPO (DClamp-PPO), which augments the PPO surrogate with a directional penalty that activates only in strict wrong-direction regions (Figure~\ref{fig:method}). We further prove that, when initialized in a strict wrong direction, DClamp-PPO moves ratios closer to $1$ than PPO (Lemma~\ref{lem:1}), and we corroborate this with experiments (Tables~\ref{tab:wrong-band} and~\ref{tab:mse}, and Figure~\ref{fig:ratio_on}).

\subsection{Importance Ratios in the Wrong Direction}
\label{subsec:IR-WD}

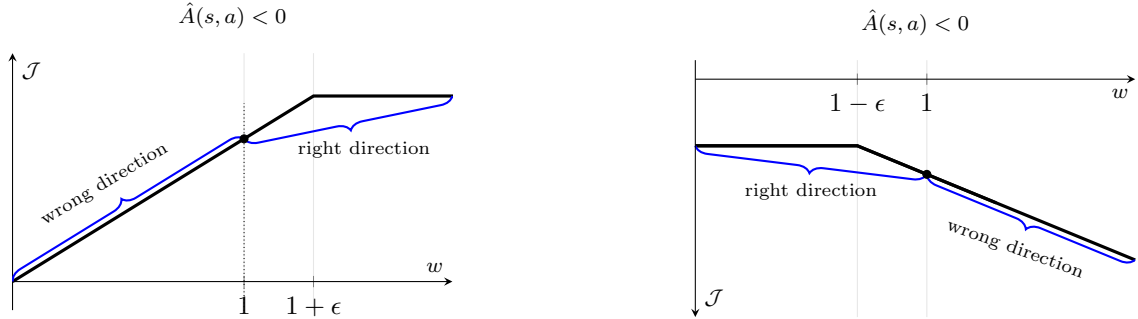
\begin{figure}[H]
\centering
\def\eps{0.3}   
\def\coef{5}    
\begin{minipage}{0.45\textwidth}

\centering
\begin{tikzpicture}
\pgfmathsetmacro{\Apos}{1}                
\pgfmathsetmacro{\Xm}{1-\eps}           
\pgfmathsetmacro{\Xp}{1+\eps}             
\pgfmathsetmacro{\YppoFlat}{\Apos*\Xp}

\pgfmathsetmacro{\YcpZero}{-(\coef-1)*\Apos*(1-\eps)}
\pgfmathsetmacro{\YcpAtXm}{\Apos*\Xm}    
\pgfplotsset{
  legend image code/.code={
    \draw[#1,line width=1.4pt] (1cm,0cm) -- (1.55cm,0cm);
  }
}

\begin{axis}[
  xmin=0, xmax=1.9, ymin=-0.2, ymax=1.6,
  axis lines=middle, axis line style={-stealth, line width=0.4pt},
  width=\textwidth, height=5cm,
  clip=true,
  grid=both, major grid style={gray!20},
  xlabel={\(w\)}, xlabel style={font=\footnotesize},
  ylabel={\(\mathcal J\)}, ylabel style={font=\footnotesize},
  title={\(\hat{A}(s,a)<0\)}, title style={font=\footnotesize},
  xtick={0,1,1+\eps}, xticklabels={$0$,$1$,$1+\epsilon$},
  ytick=\empty,
  legend style={
    font=\scriptsize,
    at={(0.02,0.98)}, anchor=north west,
    draw=none, fill=none
  }
]

\addplot[densely dotted, forget plot] coordinates {(1,-2.2) (1,1.25)};
\addplot[only marks, mark=*, mark size=1.5pt, black, forget plot]
  coordinates {(1,\Apos)};

\addplot[black, very thick] coordinates {(0,0) (\Xp,\YppoFlat) (2,\YppoFlat)};
\addlegendimage{black, very thick, mark=none}
\usetikzlibrary{decorations.pathreplacing}

\draw [decorate,decoration={brace,amplitude=5pt,mirror},blue,thick]
  (axis cs:1,1) -- (axis cs:0,0)
  node[midway,xshift=-9pt,yshift=10pt,font=\scriptsize,black, rotate=31]{wrong direction};
\draw [decorate,decoration={brace,amplitude=5pt,mirror},blue,thick]
  (axis cs:1,1) -- (axis cs:1.9,1.3)
node[midway,xshift=6pt,yshift=-12pt,font=\scriptsize,black]{right direction};

\end{axis}
\end{tikzpicture}
\end{minipage}
\hfill
\begin{minipage}{0.45\textwidth}
\centering
\begin{tikzpicture}
\pgfmathsetmacro{\Aneg}{-1}               
\pgfmathsetmacro{\Xm}{1-\eps}
\pgfmathsetmacro{\Xp}{1+\eps}
\pgfmathsetmacro{\YppoFlatN}{\Aneg*\Xm}    
\pgfmathsetmacro{\Apos}{1}

\pgfmathsetmacro{\YcpAtXpN}{\Aneg*(\Xp)}  
\pgfmathsetmacro{\YcpAtTwoN}{\coef*\Aneg*2 - ((\coef*\Aneg)-\Aneg)*(1+\eps)}

\begin{axis}[
  xmin=0, xmax=1.9, ymin=-2.5, ymax=0.2,
  axis x line=middle,
  axis y line=middle,
  axis line style={draw=none},  
  tick style={line width=0.4pt},
  clip=false,
  width=\textwidth, height=5cm,
  grid=both, major grid style={gray!20},
  title={\(\hat{A}(s,a)<0\)}, title style={font=\footnotesize},
  xtick={0,1-\eps,1},
  xticklabels={$0$,$1-\epsilon$,$1$},
  ytick=\empty,
  legend style={font=\scriptsize, at={(0.02,0.98)}, anchor=north west, draw=none, fill=none},
  after end axis/.code={
    \draw[-stealth, line width=0.4pt] (axis cs:0,0) -- (axis cs:1.9,0);
    \draw[-stealth, line width=0.4pt] (axis cs:0,0.2) -- (axis cs:0,-2.5);
    \node[font=\footnotesize, anchor=north]       at (axis cs:1.835,0) {\(w\)};
    \node[font=\footnotesize, anchor=west] at (axis cs:0,-2.3) {\(\mathcal{J}\)};
  },
]

\addplot[black, very thick] coordinates {(0,\YppoFlatN) (\Xm,\YppoFlatN) (\Xp,\Aneg*\Xp)};

\addplot[black, very thick] coordinates {(0,\YppoFlatN) (\Xm,\YppoFlatN) (1.9,\Aneg*1.9)};

\addplot[only marks, mark=*, mark size=1.5pt, black] coordinates {(1,-\Apos)};

\draw [decorate,decoration={brace,amplitude=5pt,mirror},blue,thick]
 (axis cs:1,-1) -- (axis cs:1.9,\Aneg*1.9)
node[midway,xshift=-6pt,yshift=-12pt,font=\scriptsize,black, rotate=-20]{wrong direction};
\draw [decorate,decoration={brace,amplitude=5pt,mirror},blue,thick]
  (axis cs:0,-0.7) -- (axis cs:1,-1) 
  node[midway,yshift=-12pt,font=\scriptsize,black]{right direction};

\end{axis}
\end{tikzpicture}
\end{minipage}
\caption{The right and wrong directions on the PPO surrogate function. The black dot at importance ratio 1 symbolizes the starting point at the optimization. }

\label{fig:right-wrong-direction-ppo-loss}
\end{figure}

The optimization of the surrogate in \eqref{eqn:PPO} begins with $\theta = \theta_{\text{old}}$, where the importance ratio satisfies $w_{\theta}(s,a) = \pi_{\theta}(a|s)/\pi_{\theta_{\text{old}}}(a|s) = 1$. During optimization, we say that a sample $(s,a)$ lies in the {\em wrong direction region} if $(w_{\theta}(s,a) - 1)\,\hat{A}(s,a) < 0$, meaning that the parameter update induced by $\theta$ changes the policy probability in a direction opposite to the advantage signal, i.e.,~it decreases the likelihood of advantageous actions or increases that of disadvantageous ones. The {\em right direction region} can be defined conversely as $(w_{\theta}(s,a) - 1) \hat{A}(s,a) > 0$. The right and wrong directions and their relations to the PPO objective are shown in Figure \ref{fig:right-wrong-direction-ppo-loss}.

\begin{figure}[H]
\centering

\begin{subfigure}[t]{0.24\linewidth}
  \centering
  \includegraphics[width=\linewidth]{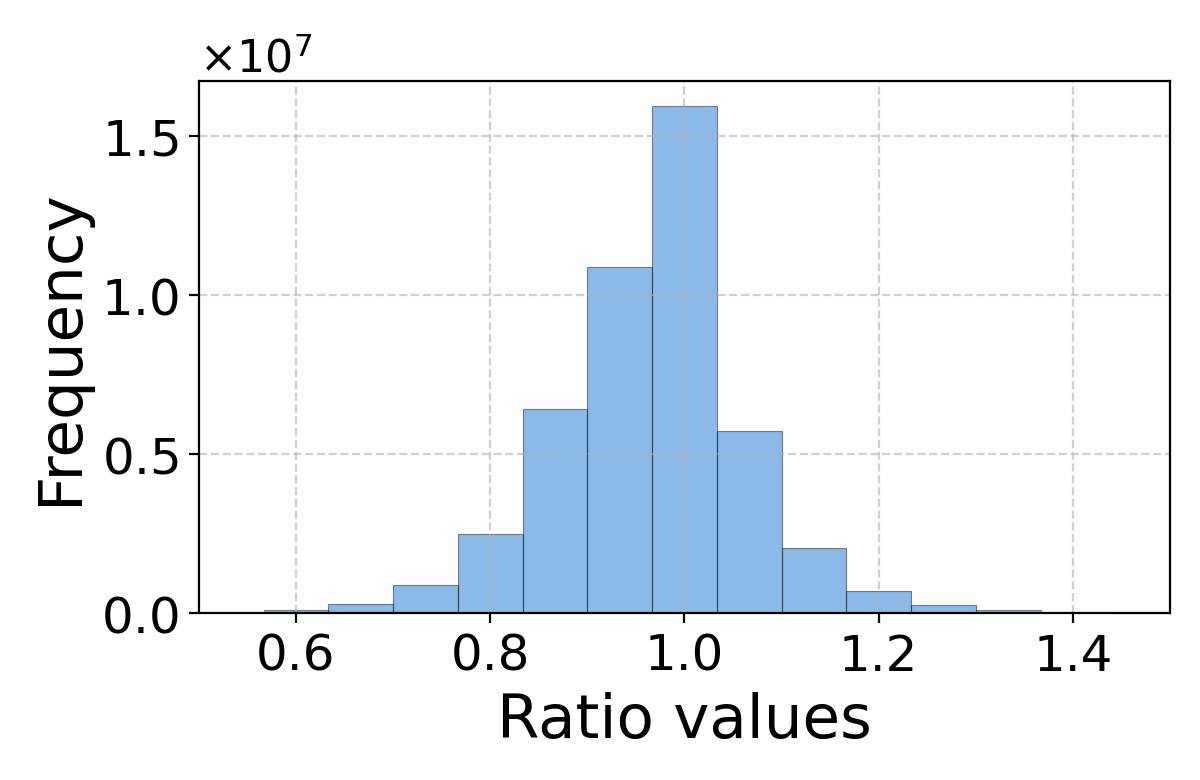}
  \caption{Ant-v4, $\hat{A}(s,a)<0$}
\end{subfigure}\hfill
\begin{subfigure}[t]{0.24\linewidth}
  \centering
  \includegraphics[width=\linewidth]{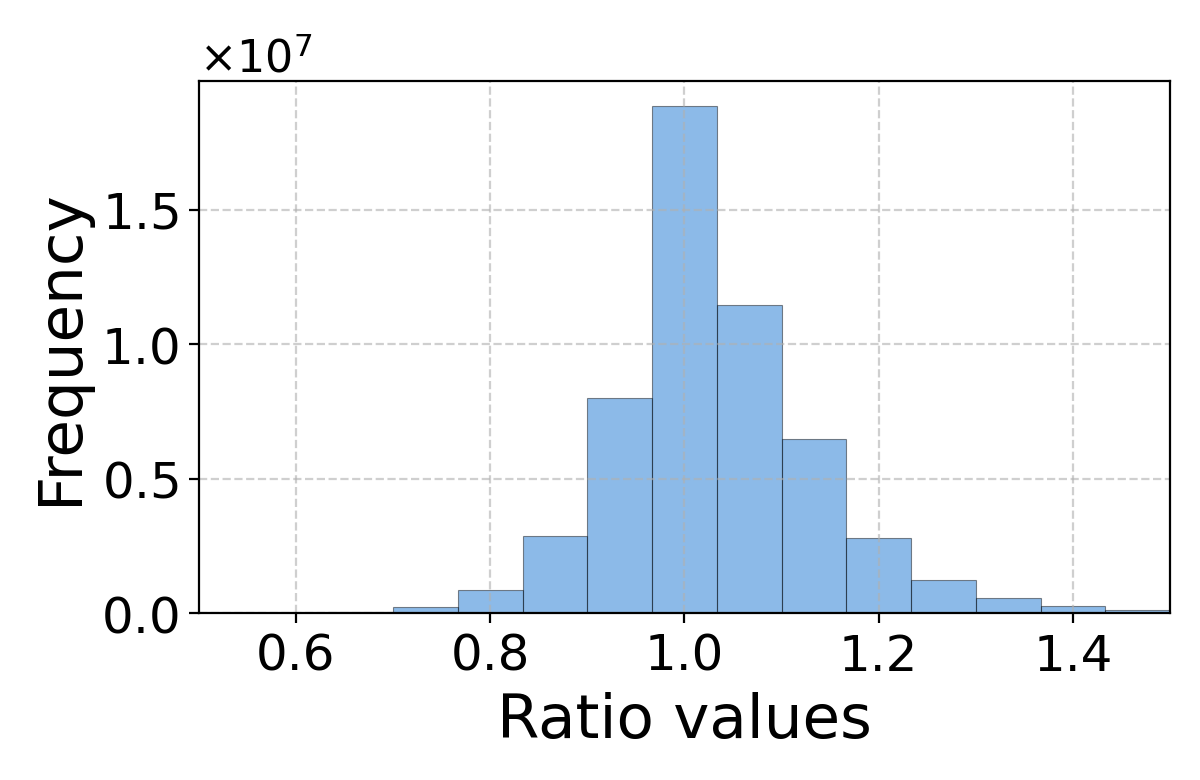}
  \caption{Ant-v4, $\hat{A}(s,a)>0$}
\end{subfigure}\hfill
\begin{subfigure}[t]{0.24\linewidth}
  \centering
  \includegraphics[width=\linewidth]{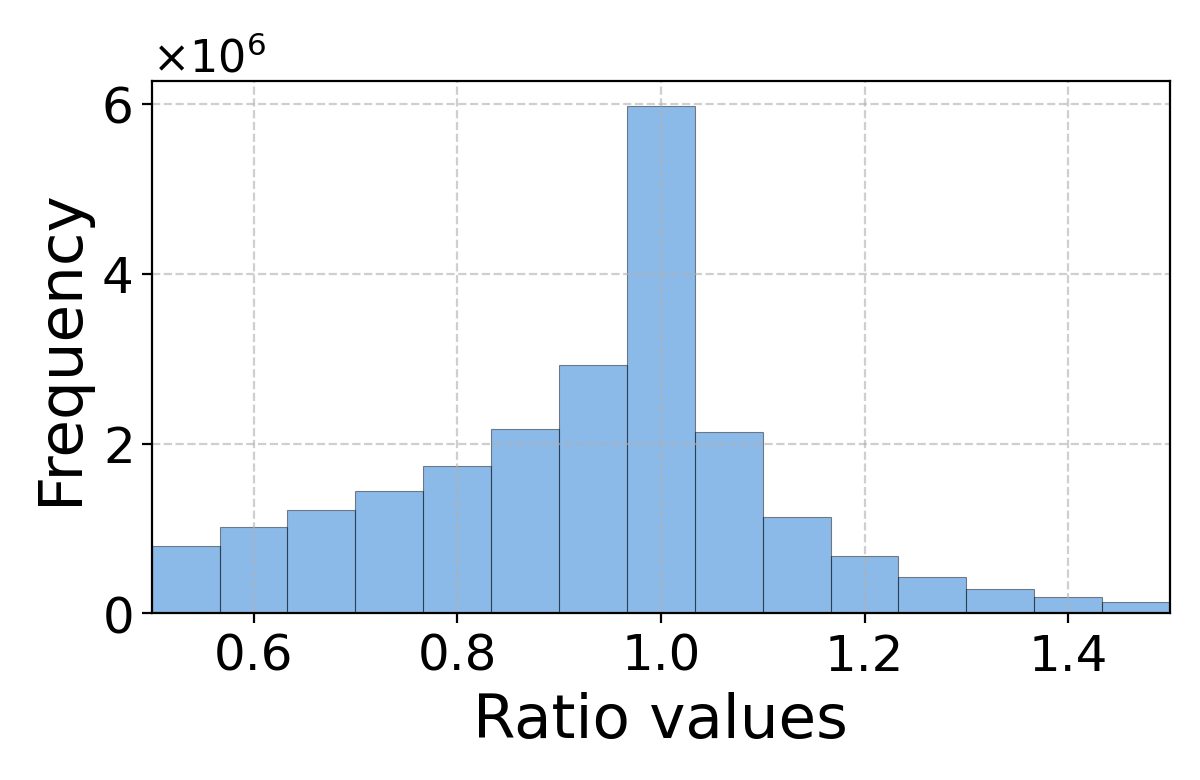}
  \caption{Hopper-v4, $\hat{A}(s,a)<0$}
\end{subfigure}\hfill
\begin{subfigure}[t]{0.24\linewidth}
  \centering
  \includegraphics[width=\linewidth]{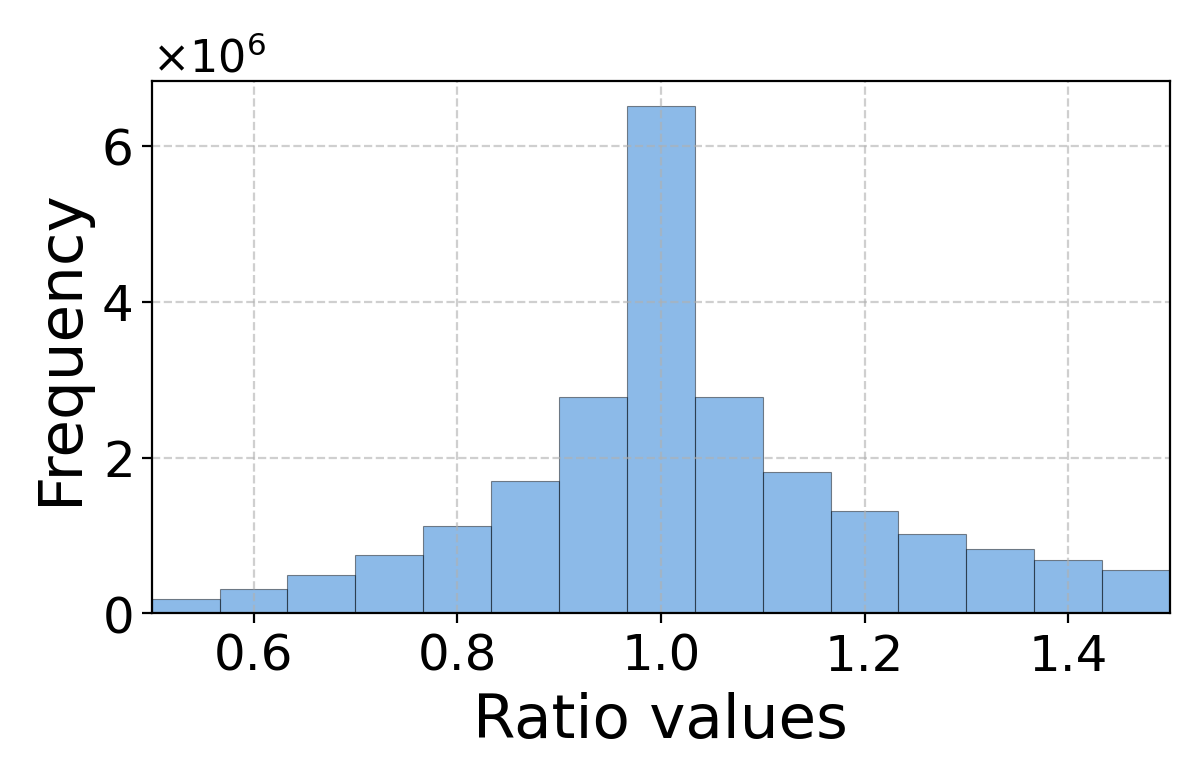}
  \caption{Hopper-v4, $\hat{A}(s,a)>0$}
\end{subfigure}

\caption{Histograms of importance ratios during PPO optimization for two MuJoCo environments.}
\label{fig:ratio_histograms}
\end{figure}

Although PPO’s objective function \eqref{eqn:PPO} is designed to discourage updates in the wrong directions, our experiments reveal that this safeguard is not always effective in practice, as illustrated in Figure~\ref{fig:ratio_histograms}. In our experiments, we observed that approximately $34.95\%$ of the ratio samples were in the wrong direction on average when the advantage was negative. Moreover, when the advantage was positive, $39.88\%$ of the ratio samples were in the wrong direction on average, with the proportion reaching as high as $49\%$ in a single environment. These ratios suggest that PPO frequently produces updates that move the policy against the advantage signal, which can undermine learning stability. Specifically, for each collected transition $(s,a)$ we computed the importance ratio $w_\theta(s,a) = \pi_{\theta}(a|s)/\pi_{\theta_{\text{old}}}(a|s)$ and the corresponding advantage estimate $\hat{A}(s,a)$. The reported percentages correspond to the empirical proportion of such samples across all updates. The histogram in Figure~\ref{fig:ratio_histograms} visualizes this distribution.

\subsection{Directional-Clamp PPO Algorithm}\label{sec:DClamp-PPO}

Motivated by the observations reported in Section~\ref{subsec:IR-WD}, we now propose our algorithm Directional-Clamp PPO (DClamp-PPO), which is mainly based on a modification to PPO clipping. Before describing DClamp-PPO, it is important to note that small deviations from $w_{\theta}(s,a)=1$ are natural during stochastic optimization and may not harm performance. Thus, to separate benign deviations from genuinely harmful ones, we introduce the notion of a {\em strict wrong direction}. More precisely, we consider an update in strictly wrong direction if the updated parameter $\theta_{\text{new}}$ assigns a probability ``significantly'' lower (higher) than the old policy to an action with positive (negative) advantage. We quantify the degree of strictness using a hyperparameter $0 \leq \beta \leq 1$ and formally define strict wrong direction as
$$
\begin{cases}
w_{\theta}(s,a) > 1 + \beta & \text{if } \; \hat{A}(s,a)<0, \\
w_{\theta}(s,a) < 1 - \beta  & \text{if } \; \hat{A}(s,a)>0.
\end{cases}
$$

DClamp-PPO steers the policy parameter $\theta$ toward the correct direction during optimization by introducing a hyperparameter that controls the strength of an additional penalty term in the surrogate objective. Importantly, this penalty is applied only when updates fall into the strict wrong direction region, thereby focusing the correction on harmful updates while tolerating benign fluctuations near $w_{\theta}(s,a)=1$. In doing so, DClamp-PPO also mitigates the overshooting problem, where probability ratios drift too far from $1$ despite clipping.

\begin{figure}[H]
\centering

\def\eps{0.3}   
\def\coef{5}   

\begin{minipage}{0.45\textwidth}
\centering
\begin{tikzpicture}
\pgfmathsetmacro{\Apos}{1}                
\pgfmathsetmacro{\Xm}{1-\eps}           
\pgfmathsetmacro{\Xp}{1+\eps}              
\pgfmathsetmacro{\YppoFlat}{\Apos*\Xp}     

\pgfmathsetmacro{\YcpZero}{-(\coef-1)*\Apos*(1-\eps)}
\pgfmathsetmacro{\YcpAtXm}{\Apos*\Xm}  

\begin{axis}[
  xmin=0, xmax=1.9, ymin=-0.2, ymax=1.6,
  axis lines=middle, axis line style={-stealth, line width=0.4pt},
  width=\textwidth, height=5cm,
  clip=true,
  grid=both, major grid style={gray!20},
  xlabel={\(w\)}, xlabel style={font=\footnotesize},
  ylabel={\(\mathcal J\)}, ylabel style={font=\footnotesize},
  title={\(\hat{A}(s,a)>0\)}, title style={font=\footnotesize},
  xtick={0,\Xm,1,2}, xticklabels={$0$,$1-\beta$,$1$,$2$},
  ytick=\empty
]
\addplot[densely dotted] coordinates {(1,-2.2) (1,1.25)};
\addplot[only marks, mark=*, mark size=1.5pt, black] coordinates {(1,\Apos)};

\addplot[black, very thick] coordinates {(0,0) (\Xp,\YppoFlat) (2,\YppoFlat)};

\usetikzlibrary{decorations.pathreplacing}

\draw [decorate,decoration={brace,amplitude=5pt,mirror},blue,thick]
  (axis cs:\Xm,\Xm) -- (axis cs:0,0)
  node[midway, xshift=5pt, yshift=17pt, font=\scriptsize, black, rotate=32]
  {strictly wrong direction};

\addplot[red, thick] coordinates {(0,\YcpZero) (\Xm,\YcpAtXm)};
\addplot[red, thick] coordinates {(\Xm,\YcpAtXm) (\Xp,\YppoFlat)};
\addplot[red, thick] coordinates {(\Xp,\YppoFlat) (2,\YppoFlat)};
\end{axis}
\end{tikzpicture}
\end{minipage}
\hfill
\begin{minipage}{0.45\textwidth}
\centering
\begin{tikzpicture}
\pgfmathsetmacro{\Aneg}{-1}            
\pgfmathsetmacro{\Xm}{1-\eps}
\pgfmathsetmacro{\Xp}{1+\eps}
\pgfmathsetmacro{\YppoFlatN}{\Aneg*\Xm}   
\pgfmathsetmacro{\Apos}{1}               
\pgfmathsetmacro{\YcpAtXpN}{\Aneg*(\Xp)}  
\pgfmathsetmacro{\YcpAtTwoN}{\coef*\Aneg*2 - ((\coef*\Aneg)-\Aneg)*(1+\eps)}

\begin{axis}[
  xmin=0, xmax=1.9, ymin=-2.5, ymax=0.2,
  axis x line=middle,
  axis y line=middle,
  axis line style={draw=none},       
  width=\textwidth, height=5cm,
  clip=true,
  grid=both, major grid style={gray!20},
  xlabel={}, ylabel={},
  title={\(\hat{A}(s,a)<0\)}, title style={font=\footnotesize},
  xtick={0,1,\Xp,2}, xticklabels={$0$,$1$,$1+\beta$,$2$},
  ytick=\empty,
  after end axis/.code={
    \draw[-stealth, line width=0.4pt] (axis cs:0,0) -- (axis cs:1.9,0); 
    \draw[-stealth, line width=0.4pt] (axis cs:0,0.25) -- (axis cs:0,-2.5); 
    \node[font=\footnotesize, anchor=north] at (axis cs:1.835,0) {\(w\)};
    \node[font=\footnotesize, anchor=west]  at (axis cs:0,-2.3)  {\(\mathcal{J}\)};
  },
]

\addplot[densely dotted] coordinates {(1,-3.6) (1,0.25)};

\addplot[black, very thick] coordinates {(0,\YppoFlatN) (\Xm,\YppoFlatN) (1.9,\Aneg*1.9)};

\addplot[red, thick] coordinates {(0,\YppoFlatN) (\Xm,\YppoFlatN)};
\addplot[red, thick] coordinates {(\Xm,\YppoFlatN) (\Xp,\Aneg*\Xp)};
\addplot[only marks, mark=*, mark size=1.5pt, black] coordinates {(1,-\Apos)};

\draw [decorate,decoration={brace,amplitude=5pt,mirror},blue,thick]
  (axis cs:\Xp,-\Xp) -- (axis cs:1.9,\Aneg*1.9)
  node[midway, xshift=-15pt, yshift=-5pt, font=\scriptsize, black, rotate=-22]
  {strictly wrong direction};

\addplot[red, thick] coordinates {(\Xp,\Aneg*\Xp) (2,\YcpAtTwoN)};
\end{axis}
\end{tikzpicture}
\end{minipage}

\caption{The surrogate objectives of PPO and DClamp-PPO, where the red curve is $\mathcal{J}_{\text{DClamp-PPO}}$ and the black curve is $\mathcal{J}_{\text{PPO}}$.}
\label{fig:method}
\end{figure}
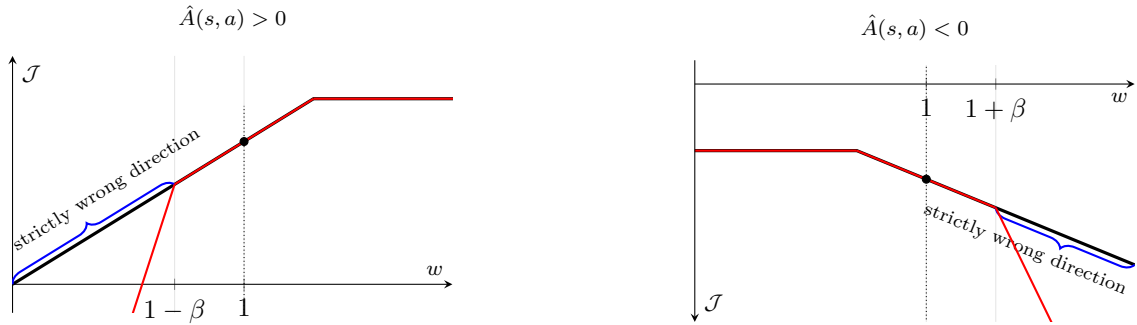

\paragraph{DClamp-PPO Surrogate}
To implement the idea described above, we introduce a hyperparameter $\alpha > 1$ that determines the slope of the penalty applied in the strictly wrong direction regions. This results in the surrogate function $\mathcal{J}_{\text{DClamp-PPO}}(\theta) $, which we define by
\begin{equation}
\label{eq:DClamp-PPO-Objective}
\hat{\mathbb{E}}_{s,a}\!\left[\min\left(w_{\theta}(s,a)\hat{A}(s,a), \,\mathrm{clip}\big(w_{\theta}(s,a),1-\epsilon,1+\epsilon\big)\hat{A}(s,a), \, 
f_{\text{DClamp}}\big(w_{\theta}(s,a),\epsilon,\alpha,\beta\big)
\hat{A}(s,a)\right)\right],
\end{equation}
where
\begin{equation*}
f_{\text{DClamp}}\big(w_{\theta}(s,a),\epsilon,\alpha,\beta\big) = 
\begin{cases}
\alpha\,  w_{\theta}(s,a)- \left(\alpha  - 1\right )\left (1-\beta \right) & \text{if } \; \hat{A}(s,a) > 0, \\
\alpha\, w_{\theta}(s,a)- \left(\alpha  - 1\right )\left(1+\beta\right) & \text{if } \; \hat{A}(s,a) < 0.
\end{cases}
\end{equation*}
Figure~\ref{fig:method} shows the difference between the surrogate functions of PPO and DClamp-PPO for both cases where the advantage estimate is positive and negative.

\noindent
Given the surrogate function~\eqref{eq:DClamp-PPO-Objective}, we can write a pseudo-code for DClamp-PPO as shown in Algorithm~\ref{alg:DClamp-PPO}.

\begin{algorithm}[H]
\caption{The pseudo-code for our Directional-Clamp PPO Algorithm.}
\label{alg:DClamp-PPO}
\begin{algorithmic}[1]
\State {\bf Initialize} $\alpha > 1, \beta \in [0, 1]$ and $\epsilon$. 
\For{iteration $=1,2,\ldots$}
    \For{actor $=1,2,\ldots,N$}
        \State Run policy $\pi_{\theta_{\text{old}}}$ in environment for $T$ time-steps.
        \State Compute generalized advantage estimators $\hat{A}(s_1,a_1), \ldots, \hat{A}(s_T,a_T)$.
    \EndFor
    \For{epoch $=1,2,\ldots,K$}
        \State Optimize policy $\pi_{\theta}$ based on $\mathcal{J}_{\text{DClamp-PPO}}$ in~\eqref{eq:DClamp-PPO-Objective} using the coefficients $\alpha$ and $\beta$.
    \EndFor
    \State $\theta_{\text{old}} \gets \theta$
\EndFor
\end{algorithmic}
\end{algorithm}

\begin{figure}[H]
\centering

\begin{subfigure}[t]{0.24\linewidth}
  \centering
\includegraphics[width=\linewidth,height=0.34\textheight,keepaspectratio]{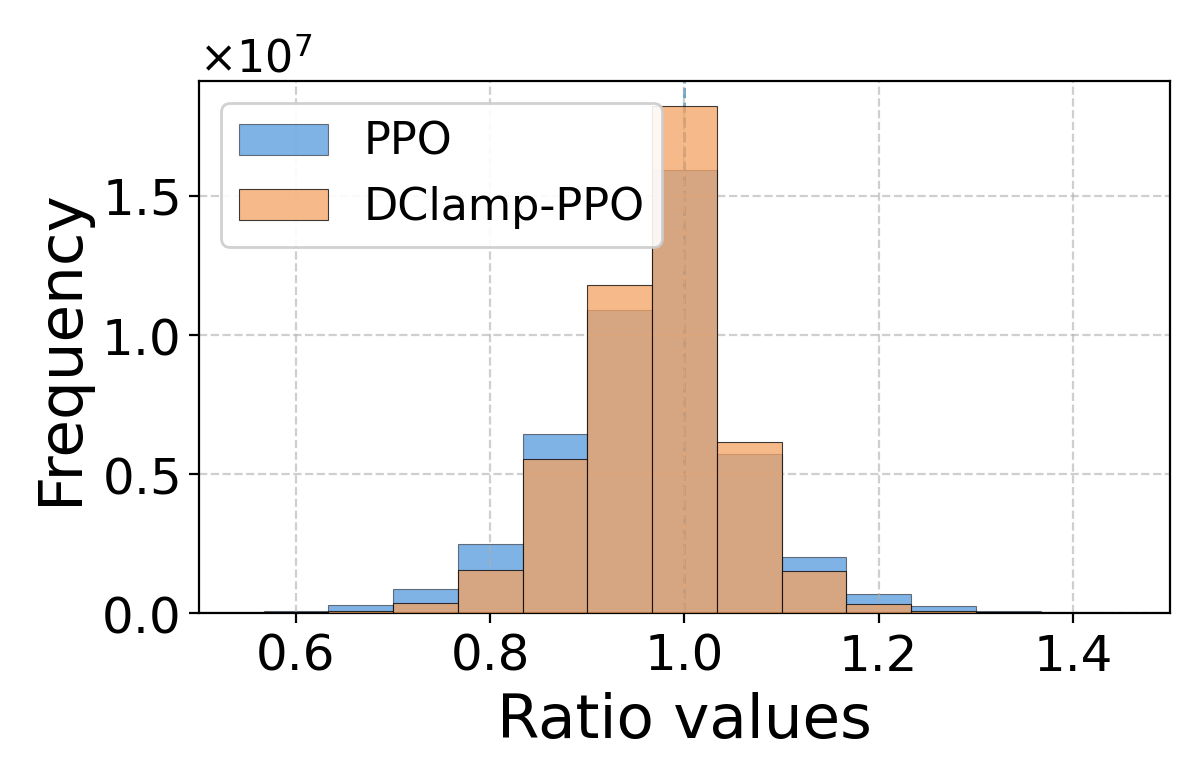}
  \caption{Ant-v4, $\hat{A}(s,a)<0$}
\end{subfigure}%
\hfill
\begin{subfigure}[t]{0.24\linewidth}
  \centering
\includegraphics[width=\linewidth,height=0.34\textheight,keepaspectratio]{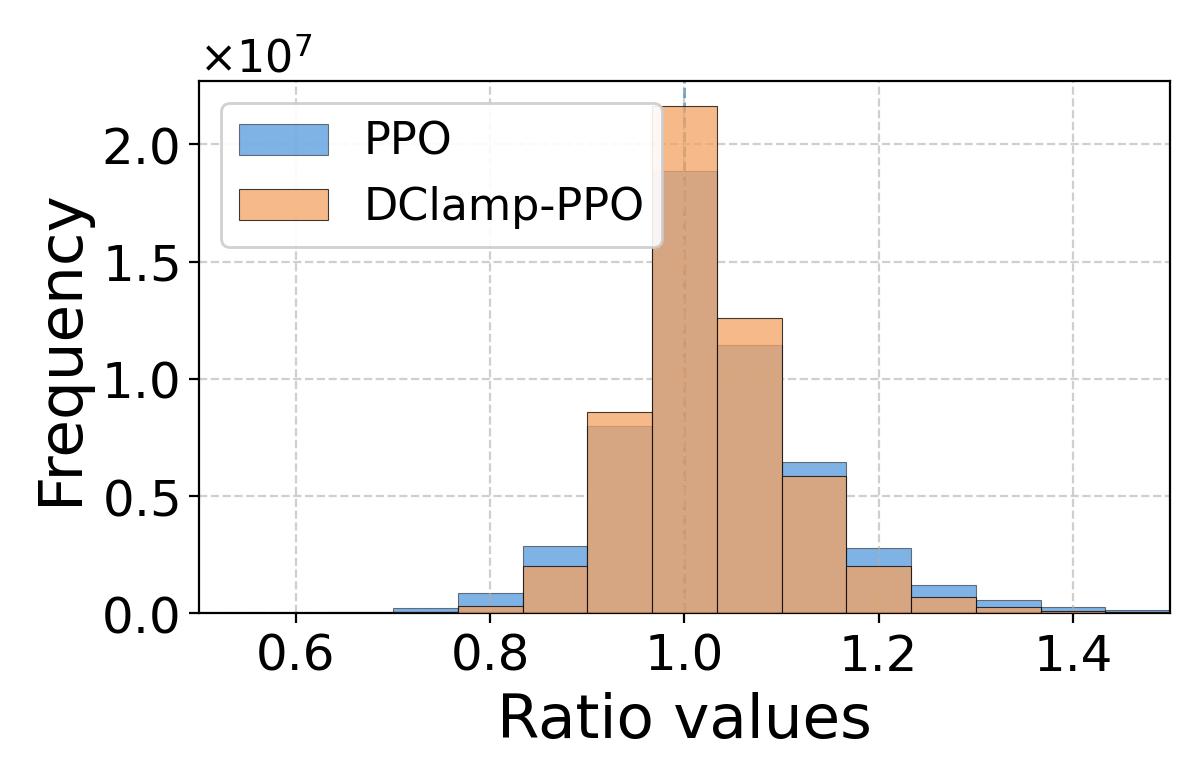}
  \caption{Ant-v4, $\hat{A}(s,a)>0$}
\end{subfigure}%
\hfill
\begin{subfigure}[t]{0.24\linewidth}
  \centering
\includegraphics[width=\linewidth,height=0.34\textheight,keepaspectratio]{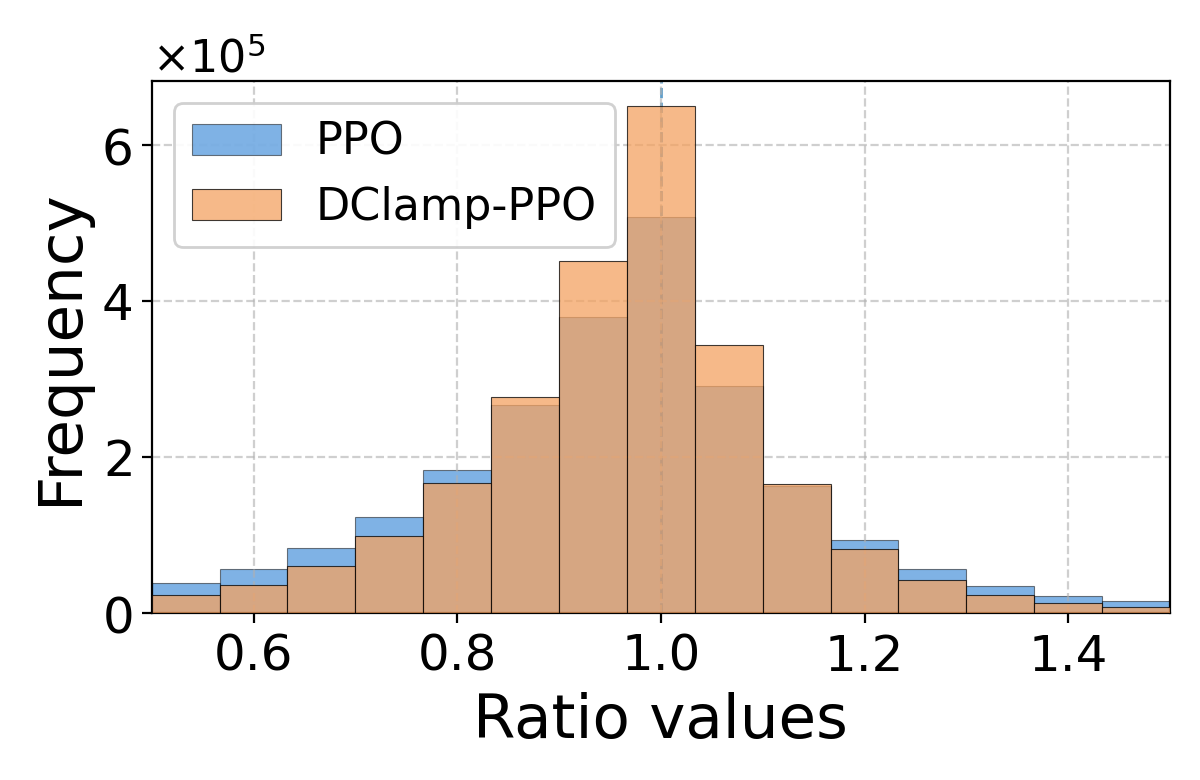}
  \caption{Hopper-v4, $\hat{A}(s,a)<0$}
\end{subfigure}%
\hfill
\begin{subfigure}[t]{0.24\linewidth}
  \centering
\includegraphics[width=\linewidth,height=0.34\textheight,keepaspectratio]{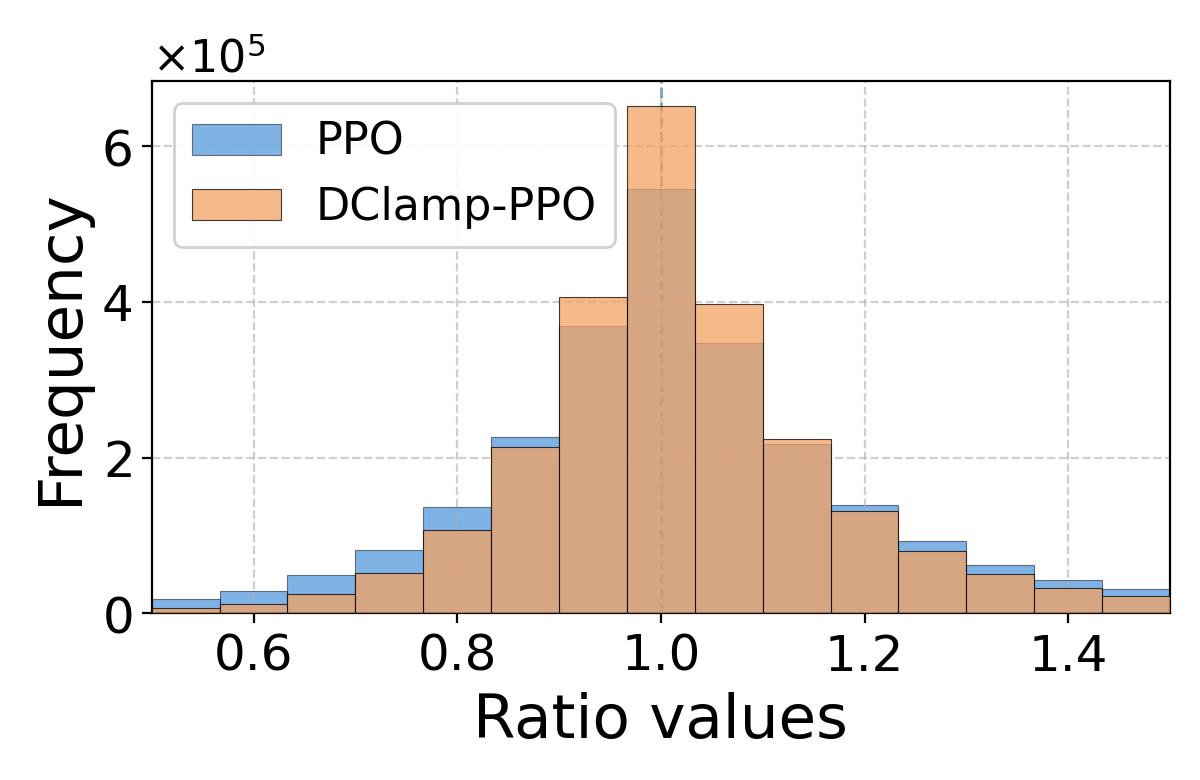}
    \caption{Hopper-v4, $\hat{A}(s,a)>0$}
\end{subfigure}
\caption{Histograms of policy ratio values $r_{\theta}(s,a)$ measured during training for DClamp-PPO and PPO on representative MuJoCo tasks. Each histogram aggregates tens of millions of samples collected across optimization steps.}
\label{fig:ratio_on}
\end{figure}

With the optimal values of hyper-parameters $\alpha$ and $\beta$ (see Appendix~\ref{app:hyperparams}), DClamp-PPO substantially reduces the proportion of ratio samples in the strict wrong direction region, for both positive and negative advantages, across all the MuJoCo environments in our experiments (see Table~\ref{tab:wrong-band} and Figure~\ref{fig:ratio_on}). The results in Table~\ref{tab:wrong-band} indicate that the average proportion of ratio samples in the strict wrong direction regions across seven MuJoCo environments for DClamp-PPO is roughly two thirds of that for PPO. Figure~\ref{fig:ratio_on} also shows that DClamp-PPO substantially reduces the number of ratio samples falling into the strict wrong direction region compared to PPO. The fact that most ratios remain concentrated around $w_{\theta}(s,a)=1$ indicates that updates stay closer to the intended trust-region under DClamp-PPO than PPO. These results demonstrate that the proposed DClamp-PPO objective mitigates the harmful update phenomenon, identified earlier in PPO, and steers the policy in the desired direction.

\begin{table}[H]
\centering
\caption{Proportion of strict wrong directions evaluated across MuJoCo environments.}
\label{tab:wrong-band}
\begin{tabular}{|l|cc|cc|}
\hline
\multirow{2}{*}{Environment} 
& \multicolumn{2}{c|}{$\hat{A}(s,a)<0$} 
& \multicolumn{2}{c|}{$\hat{A}(s,a)>0$} \\
& DClamp-PPO & PPO & DClamp-PPO & PPO \\
\hline
HalfCheetah-v4 & \textbf{0.0647} & 0.1014 & \textbf{0.0745} & 0.1258 \\
Hopper-v4      & \textbf{0.0568} & 0.0883 & \textbf{0.0600} & 0.1062 \\
Swimmer-v4     & \textbf{0.0153} & 0.0388 & \textbf{0.0110} & 0.0231 \\
Walker2d-v4    & \textbf{0.0293} & 0.0461 & \textbf{0.0284} & 0.0458 \\
Ant-v4         & \textbf{0.0428} & 0.0692 & \textbf{0.0445} & 0.0753 \\
Humanoid-v4    & \textbf{0.0342} & 0.0406 & \textbf{0.0371} & 0.0471 \\
Reacher-v4     & \textbf{0.0428} & 0.0671 & \textbf{0.0278} & 0.0518 \\
\hline
\textbf{Overall} 
& \textbf{0.0405} & 0.0617 
& \textbf{0.0425} & 0.0689 \\
\hline
\end{tabular}
\end{table}

We can also formally justify the above numerical results using the following finding. Given a horizon $T \in \mathbb{N}$ and a parameterized policy $\pi_\theta$ continuous with respect to $\theta$, we define the set of all iterations $t\in[T]$ in the mini-batch for which the ratio $w_{\theta_0}(s_t,a_t)$ is in the strict wrong direction as $\Omega_{T} = \big\{t\in [T] : |w_{\theta_0}(s_t,a_t) - 1| > \beta \; \wedge \; (w_{\theta_0}(s_t,a_t) - 1)\hat A(s_t,a_t) < 0 \big\}$. For simplicity, we use the following two notations, $\theta_1^\text{PPO} = \theta_{0}+\gamma\nabla \mathcal{J}_\text{PPO}\!\left(\theta_{0}\right)$ and $\theta_1^\text{DClamp-PPO} = \theta_{0}+\gamma\nabla \mathcal{J}_\text{DClamp-PPO}\!\left(\theta_{0}\right)$, for any given policy parameter $\theta_0$. 
\begin{lemma}[DClamp-PPO update moves the ratio toward $1$ in strict wrong direction]
\label{lem:1}
For any $t\in \Omega_{T}$, if $w_{\theta_0}(s_t,a_t)$ satisfies $\sum_{t'\in\Omega_T} \langle\nabla w_{\theta_0}(s_t,a_t),\nabla w_{\theta_0}(s_{t'},a_{t'})\rangle \hat A(s_t,a_t)\hat A(s_{t'},a_{t'})>0$, then there exists $\bar{\gamma}>0$, such that for any $\gamma \in (0,\bar{\gamma})$, we have 
$$
|w_{\theta_1^\text{DClamp-PPO}}(s_t,a_t)-1|^2<|w_{\theta_1^\text{PPO}}(s_t,a_t)-1|^2.
$$
\end{lemma}
Lemma \ref{lem:1} shows that when starting from a point in the strict wrong direction region, DClamp-PPO updates move the ratio closer to 1 than PPO. We empirically verify this behavior in Table \ref{tab:mse}. The results in this table show that DClamp-PPO achieves a lower MSE of ratios from $1$ compared to PPO for all $7$ MuJoCo environments (except a single case when $\hat{A}(s,a) > 0$). This suggests that our proposed surrogate objective better constrains updates within the trust-region, leading to more stable policy changes, whereas PPO exhibits greater variability in its ratio distributions.

\begin{table}[H]
\centering
\caption{Mean squared error (MSE) from 1 of importance sampling ratios across MuJoCo games.}
\label{tab:mse}
\begin{tabular}{|l|cc|cc|}
\hline
\multirow{2}{*}{Environment} 
& \multicolumn{2}{c|}{$\hat{A}(s,a)<0$} 
& \multicolumn{2}{c|}{$\hat{A}(s,a)>0$} \\
& DClamp-PPO & PPO & DClamp-PPO & PPO \\
\hline
HalfCheetah-v4 & \textbf{0.0134} & 0.0212 & \textbf{0.0155} & 0.0250 \\
Hopper-v4      & \textbf{0.0369} & 0.0716 & \textbf{0.0757} & 0.1510 \\
Swimmer-v4     & \textbf{0.0072} & 0.0133 & \textbf{0.0062} & 0.0099 \\
Ant-v4         & \textbf{0.0070} & 0.0110 & \textbf{0.0080} & 0.0133 \\
Reacher-v4     & \textbf{0.0365} & 0.0686 & \textbf{0.0308} & 0.0534 \\
Humanoid-v4    & \textbf{0.0583} & 0.0720 & 1.5816 & \textbf{1.3608} \\
Walker2d-v4    & \textbf{0.0058} & 0.0080 & \textbf{0.0059} & 0.0083 \\
\hline
\end{tabular}
\label{fig:redmse}
\end{table}

\section{Related Work} \label{sec:related-work}
    Proximal Policy Optimization (PPO) implementation requires careful attention to numerous technical details and optimization strategies that significantly impact performance outcomes \citep{huang202237}. The algorithm's distinctive clipped objective function demonstrates superior performance compared to vanilla policy gradient methods and traditional actor-critic approaches \citep{andrychowicz2021matters}, while achieving comparable results, with less computational cost, to Trust Region Policy Optimization (TRPO) when hyperparameters are appropriately tuned \citep{engstrom2019implementation}. There are a lot of previous attempts to further improve PPO. Most of these approaches focus on modifying the objective's behavior in the right direction, where the new policy decreases/increases the probability of under-/over-performing actions.
\medskip

    One of these works, namely Leaky PPO~\citep{han2024leaky}, points out two main deficiencies rooted in the clipped PPO objective function. The first is the loss of gradient information outside the clipping band. The second is the pessimistic estimate: while TRPO’s divergence-based constraint is averaged over the action space, without guaranteeing that each action component remains below the threshold, clipped PPO transforms this into a ratio-based constraint applied to every individual action. Building on these two limitations, the authors proposed Leaky PPO, inspired by the Leaky ReLU activation function. The objective allows gradients to flow beyond the clipping band, with the amount of leakage controlled by a hyperparameter $\alpha$. Since the optimal value of $\alpha$ may not be known a priori, they further introduced Para-PPO, which learns $\alpha$ automatically, though it was found to be performing inferior to Leaky PPO. Overall, Leaky PPO has been shown to achieve state-of-the-art performance.
\medskip

    Another line of work considered a different direction.~\cite{wang2020truly} observed that PPO does not strictly restrict the likelihood ratio, which is supposed to be the core idea of the algorithm. They further showed that PPO cannot theoretically enforce a true trust-region constraint. To address this, they proposed PPO-RB, which introduces a new clipping function together with a rollback operation. The rollback, controlled by a hyperparameter $\alpha$, determines the strength of the correction and is proven to better confine the likelihood ratio. Empirically, PPO-RB demonstrated strong performance compared to PPO across MuJoCo tasks. Their final algorithm, Truly-PPO, combines clipping based on trust regions where the triggering condition based on PPO ratios is replaced by a criterion based on trust regions – with a rollback operation on KL divergence. 
\medskip

    Several papers investigate the theoretical guarantees of PPO’s optimization procedure. Notably, \cite{huang2024ppo} analyzed PPO’s convergence properties, while \cite{jin2024stationary} established conditions under which PPO converges to stationary points. These works provide valuable insights into the limitations of PPO from a theoretical standpoint, complementing our empirical analysis of wrong-direction updates.
\medskip

    Beyond theory, recent work has focused on improving PPO’s surrogate loss and its treatment of the trust-region. SimPO \citep{meng2024simpo}, for example, strengthens the trust-region constraint by replacing the KL-divergence with a quadratic penalty. From our perspective, this simultaneously addresses both right and wrong direction updates, whereas our method explicitly targets wrong-direction samples. Other approaches include penalties based on corresponding induced metrics, which aim to provide a more principled replacement for the KL-divergence. \cite{garg2021proximal} study the heavy-tailed nature of policy gradient distributions, showing how PPO can become increasingly off-policy relative to $\pi_{\text{old}}$ during optimization. Adaptive trust-region methods have also been proposed, such as \cite{wang2019trust}, which dynamically adjust the range of allowable updates to improve exploration.

\section{Experiments}
    To evaluate the effectiveness of our approach, we compare DClamp-PPO against PPO and its variants -- PPO-RB~\citep{wang2020truly} and Leaky PPO~\citep{han2024leaky}. These two variants modify the objective function in the right direction, showing strong empirical performance, with Leaky PPO reported to achieve state-of-the-art results. Extensive experiments are conducted on MuJoCo environments, and the hyperparameter settings and implementation details provided in Appendix~\ref{app:hyperparams}.
\medskip

\begin{figure}[H]
\centering
\begin{minipage}[t]{0.48\linewidth}
  \centering
  \includegraphics[width=\linewidth, keepaspectratio]{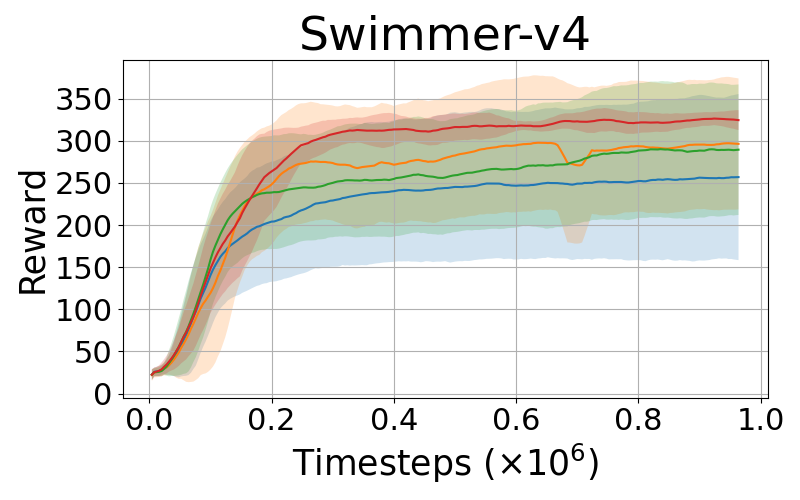}
\end{minipage}
\hfill
\begin{minipage}[t]{0.48\linewidth}
  \centering
  \includegraphics[width=\linewidth, keepaspectratio]{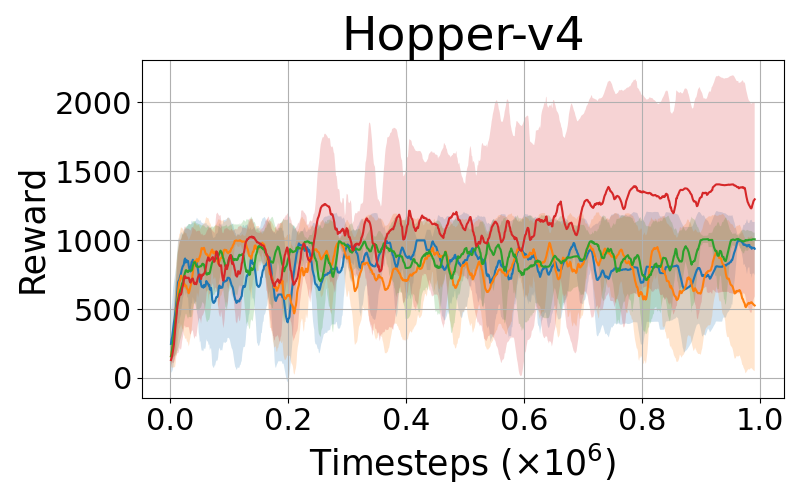}
\end{minipage}

\vspace{0.5em} 
\begin{minipage}[t]{0.48\linewidth}
  \centering
  \includegraphics[width=\linewidth, keepaspectratio]{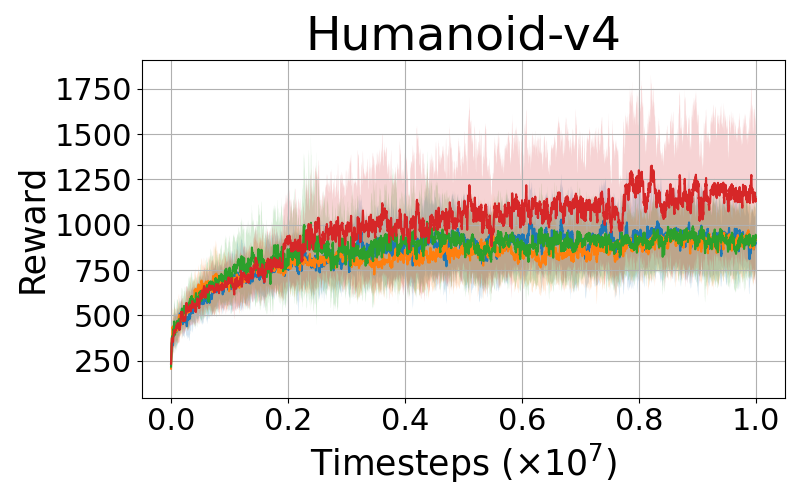}
\end{minipage}
\hfill
\begin{minipage}[t]{0.48\linewidth}
  \centering
  \includegraphics[width=\linewidth, keepaspectratio]{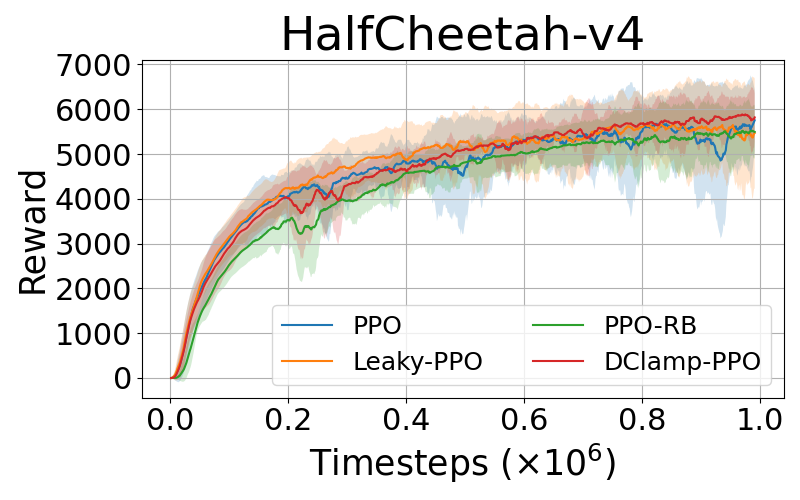}
\end{minipage}

\caption{Performance comparisons of PPO, PPO-Leaky, PPO-RB, and DClamp-PPO (ours) across MuJoCo environments.}

\label{fig:all_envs}
\end{figure}

    The training trends across environments are illustrated in Figure \ref{fig:all_envs}. Each curve represents the average episodic return as a function of training progress, where an episode denotes one complete rollout of the agent until termination. For every 1,000 episodes, the model is evaluated over 10 independent evaluation episodes, and the mean return across these evaluations is recorded. All results are averaged over 5 random seeds, and the shaded regions indicate $\pm$ standard deviation. The curves are further smoothed using a moving-average window of 10 to reduce short-term variance. We observe that in Swimmer-v4, Hopper-v4, and Humanoid-v4 environments, the proposed DClamp-PPO achieves steeper learning curves during training and converges to significantly superior performance compared to the baseline methods. In HalfCheetah-v4, DClamp-PPO exhibits similar training behavior to PPO but still converges to superior performance, though the improvement is less pronounced.

\begin{table}[H]
\centering
Averaged last 10 evaluation rewards \\[0.3em]
\hrule
\vspace{0.3cm}
\begin{tabular}{|l|c|c|c|c|c|}
\hline
Tasks          & PPO       & PPO-RB    & Leaky PPO & DClamp-PPO     & \% Change (vs PPO) \\ \hline
HalfCheetah-v4 & 5753.40   & 5494.25   & 5468.16   & \textbf{5816.94}   & 1.1\%   \\
Hopper-v4      & 937.63    & 1006.24   & 525.76    & \textbf{1296.55}   & 38.3\%  \\
Swimmer-v4     & 256.83    & 289.36    & 296.24    & \textbf{324.42}    & 26.3\%  \\
Ant-v4         & 5813.77   & \textbf{5969.68}   & 5973.11   & 5763.29   & -0.9\%   \\
Reacher-v4     & -4.78     & -4.67     & -4.72     & \textbf{-4.66}     & 2.7\%   \\
Humanoid-v4    & 900.60    & 929.90    & 914.12    & \textbf{1137.30}   & 26.3\%  \\
Walker2d-v4    & 3750.60   & 3078.21   & \textbf{3842.29}   & 3765.47   & 0.4\%   \\ \hline
\end{tabular}
\caption{Comparison of last-10 average rewards across MuJoCo tasks. 
Bold indicates the best result per environment. 
The percentage change is computed relative to PPO.}
\label{tab:last10}
\end{table}

\begin{table}[H]
\centering
Averaged top 10 episode rewards \\[0.3em]
\hrule
\vspace{0.3cm}
\begin{tabular}{|l|c|c|c|c|c|}
\hline
Tasks          & PPO       & PPO-RB    & Leaky PPO & DClamp-PPO & \% Change (vs PPO) \\ \hline
HalfCheetah-v4 & 5945.76   & 5617.86   & 5840.40   & \textbf{5951.39} & 0.09\%  \\
Hopper-v4      & 1017.51   & 1016.92   & 1009.07   & \textbf{1413.65} & 38.90\% \\
Swimmer-v4     & 259.19    & 294.28    & 300.20    & \textbf{327.82}  & 26.49\% \\
Ant-v4         & 6596.15   & \textbf{6738.53}   & 6560.68   & 6673.82   & 1.18\%  \\
Reacher-v4     & \textbf{-3.31}   & -3.44     & -3.66    & -3.42    & -3.38\% \\
Humanoid-v4    & 1119.66   & 1244.01   & 1054.83   & \textbf{1486.52} & 32.80\% \\
Walker2d-v4    & 4024.42   & 3953.52   & 4214.24   & \textbf{4294.54} & 6.72\%  \\ \hline
\end{tabular}
\caption{Comparison of averaged top-10 episode rewards across MuJoCo tasks. 
Bold indicates the best result per environment. 
The \% Change column reports the relative improvement of DClamp-PPO over PPO.}
\label{tab:rewards}
\end{table}

Tables \ref{tab:last10} and \ref{tab:rewards} provide a quantitative comparison of performance across seven MuJoCo environments.
Table \ref{tab:last10} reports the average return over the last 10 evaluations of training, reflecting the final stability and convergence level of each method.
Table \ref{tab:rewards} instead reports the average of the top 10 evaluation rewards, capturing the best performance achieved during training. These complementary metrics are included to evaluate not only the eventual performance after convergence but also the peak capability each algorithm can reach during optimization. Across both measures, DClamp-PPO consistently matches or surpasses baseline methods—PPO, Leaky-PPO, and PPO-RB—in most environments. The improvements are particularly notable in Hopper-v4, Swimmer-v4, and Humanoid-v4, where DClamp-PPO achieves up to nearly 40\% higher returns compared to PPO. Even in environments where it does not attain the best score (e.g., Ant-v4, Reacher-v4), the difference remains small, indicating robust and competitive performance. Overall, these results demonstrate that incorporating directional penalties into PPO yields superior or comparable outcomes across diverse continuous-control tasks, highlighting the general effectiveness and stability of DClamp-PPO.
\medskip

The reduction in strictly wrong-direction updates, as shown in Table~\ref{tab:wrong-band}, correlates strongly with these performance improvements, particularly in environments where PPO exhibited a high baseline rate of such harmful updates (e.g., Hopper-v4 and Humanoid-v4). However, in Ant-v4 and Reacher-v4, despite the decrease in strict wrong-direction ratios, performance remains comparable to PPO. This suggests that, while suppressing harmful updates generally enhances learning stability and efficiency, overall returns also depend on environment-specific factors such as reward scaling, exploration difficulty, and the influence of mild ratio deviations.
\medskip

Our analysis of hyperparameters provides additional insights. We find that tuning $\beta$ may not decrease the overall proportion of wrong-direction samples compared to PPO; in fact, with the optimal $\beta$, it slightly increases on average (see Table~\ref{tab:wrong-band-comparison} in Appendix~\ref{App:addExp}). Reducing the value of $\beta$ generally lowers the proportion of wrong-direction samples but results in worse performance across environments compared with the optimal $\beta$ (see Figure~\ref{fig:diffbeta} in Appendix~\ref{App:addExp}). A plausible explanation is that if $\beta$ is too small, the new policy diverges too far from the old one, causing instability; moreover, occasional wrong-direction updates may actually be beneficial for learning, by promoting exploration and preventing premature policy collapse. The value of $\alpha$ also plays a critical role: smaller values impose only weak penalties and yield inferior performance, while excessively large values can destabilize learning (see Figure~\ref{fig:diffalpha} in Appendix~\ref{App:addExp}). Striking the right balance in $\alpha$ is therefore crucial for the stability and effectiveness of DClamp-PPO.

\section{Conclusion}

In this paper, we identified a fundamental limitation of PPO: although its objective is designed to steer policy updates in the direction of the advantage, there is still a significant proportion of action probabilities that fail to do so in practice. 

The proposed DClamp-PPO explicitly addresses this issue by penalizing strictly wrong-direction updates through a modified surrogate objective that introduces directional penalties only when updates deviate significantly from the intended direction. We empirically show, that this penalty can substantially reduce the frequency of harmful updates across environments, confirming that DClamp-PPO effectively suppresses wrong-direction policy changes while preserving beneficial exploration. The observed reduction in the strictly wrong-direction updates generally translates to improved empirical performance. We showed that DClamp-PPO achieves substantial improvements across multiple MuJoCo environments. Our theoretical analysis shows that DClamp-PPO produces ratio updates closer to 1 than standard PPO when initialized in strict wrong-direction regions, thereby maintaining importance ratios within the trust-region more effectively. This theoretical insight provides an additional justification of the proposed method from the angle of trust-region. 
\medskip

The effectiveness of DClamp-PPO highlights an important research direction of focusing on the ``wrong direction'' regions, which have been largely overlooked in the literature. Future work could explore different penalty shapes to effectively penalize updates in the ``wrong direction'' and investigate how to adaptively adjust these penalties based on training dynamics.

\newpage

\bibliography{references}
\bibliographystyle{iclr2026_conference}  

\newpage

\appendix
\section{Proofs}
\begin{proof}[Proof of \cref{lem:1}]
    Since $t \in \Omega_T$, we either have that $w_{\theta_0}(s_t,a_t) < 1 - \beta$ and $\hat A(s_t,a_t) > 0$ or $w_{\theta_0}(s_t,a_t) > 1 + \beta$ and $\hat A(s_t,a_t) < 0$. We will show the result for the first case, as the second case is analogous. For simplicity of the proof, we define
    \begin{equation*}
    \phi\left(\gamma\right) = w_{\!\theta_{0}+\gamma\nabla\mathcal{J}_\text{DClamp-PPO}\!\left(\theta_{0}\right)}(s_t,a_t) - w_{\!\theta_{0}+\gamma\nabla\mathcal{J}_\text{PPO}\!\left(\theta_{0}\right)}(s_t,a_t).
    \end{equation*}
    Then, by computing its gradient, we get
    \begin{align*}
        \phi'\left(0\right) = \nabla w_{\theta_0}(s_t,a_t)^{\top}\!\left(\nabla \mathcal{J}_\text{DClamp-PPO}\!\left(\theta_{0}\right) - \nabla \mathcal{J}_\text{PPO}\!\left(\theta_{0}\right)\right) \\= (\alpha-1)\frac{1}{|\Omega_T|}\sum_{t'\in\Omega_T} \langle\nabla w_{\theta_0}(s_t,a_t),\nabla w_{\theta_0}(s_{t'},a_{t'})\rangle \hat A(s_{t'},a_{t'})
    \end{align*}
    where the equality follows from the definition of $\mathcal{J}_\text{DClamp-PPO}$. Since $\alpha > 1$ and the assumption, we get $\phi'\left(0\right) > 0$. Hence, there exists $\bar{\gamma} > 0$ such that for all $\gamma \in (0, \bar{\gamma})$, we have $\phi(\gamma) > \phi(0)$. This implies that
    \begin{equation} \label{L:Tech:1}
    w_{\theta_1^\text{PPO}}(s_t,a_t)<w_{\theta_1^\text{DClamp-PPO}}(s_t,a_t).
    \end{equation}
    Moreover, from the continuity of $w_{\theta}(s_t,a_t)$ in $\theta$ there exists $\bar{\delta} > 0$ such that for all $\delta \in (0 , \bar{\delta})$ we have $|w_{\theta_0 + \delta\hat{u}}(s_t,a_t) - w_{\theta_0}(s_t,a_t)| < 1 - w_{\theta_0}(s_t,a_t)$ for $\hat{u} = \frac{\nabla  \mathcal{J}_\text{DClamp-PPO}(\theta_0)}{\|\nabla \mathcal{J}_\text{DClamp-PPO}(\theta_0)\|}$\footnote{$\hat u$ is well defined by the assumption.}. This implies that $w_{\theta_0 + \delta\hat{u}}(s_t,a_t)< 1$. For all $\gamma\in \left(0,\min\left(\bar \gamma,\delta\|\nabla \mathcal{J}_\text{DClamp-PPO}(\theta_0)\|\right)\right)$, by combining with \eqref{L:Tech:1} yields that $w_{\theta_1^\text{PPO}}(s_t,a_t) < w_{\theta_1^\text{DClamp-PPO}}(s_t,a_t) < 1$. Therefore
    \begin{equation*}
        |w_{\theta_1^\text{DClamp-PPO}}(s_t,a_t)-1|^2< |w_{\theta_1^\text{PPO}}(s_t,a_t)-1|^2,
    \end{equation*}
    which proves the desired result.
\end{proof}
\newpage
\section{PPO Hyperparameter Configurations} \label{app:hyperparams}
All our experiments were implemented using the open-source Stable-Baselines3 library \citep{raffin2021stable}, with hyperparameter configurations adapted from the RL Baselines3 Zoo repository \citep{rl-zoo3}. We report them here.
\medskip

The authors of Leaky-PPO and PPO-RB provide the optimal coefficients in the appendices of their respective works~\citep{han2024leaky, wang2020truly}.
\medskip

Before reporting the results, we clarify several experimental details and definitions used in Tables~\ref{tab:last10}–\ref{tab:rewards}. The flag Normalize = True indicates that both observations and rewards are normalized during training. The parameter $n_{\text{envs}}$ denotes the number of parallel environments used during training. The MlpPolicy architecture refers to a multilayer perceptron policy and value network implemented in Stable-Baselines3, consisting of two hidden layers with 64 units each and \texttt{tanh} activations. The variable $n_{\text{timesteps}}$ specifies the total number of environment interactions during training, while $n_{\text{steps}}$ defines the number of rollout steps collected per environment before each policy update. Each policy update uses mini-batches of the specified batch size and is optimized for $n_{\text{epochs}}$ gradient epochs per update.
\medskip

Other hyperparameters follow the standard PPO setup. The final objective of PPO \citep{schulman2017proximal} combines the clipped policy surrogate with a value function and an entropy bonus. Formally, the overall objective is
\begin{align}
\mathcal{J}(\theta,\phi)
&=
\mathbb{\hat E}_{s,a}\!\left[
\mathcal J_{\text{PPO}}(\theta)
\;-\;
c_{\text{vf}}\,\mathcal J_{\text{vf}}(\phi)
\;+\;
c_{\text{ent}}\,\mathcal J_{\text{ent}}(\theta)
\right],
\end{align}
where $\mathcal J_{\text{vf}}(\phi) = (V_\phi(s_t) - V_t^{\text{targ}})^2$ is a squared-error loss and $\mathcal J_{\text{ent}}(\theta)$ denotes the entropy bonus. The coefficients $c_{\text{vf}}$ and $c_{\text{ent}}$ balance the critic and entropy contributions. During optimization, gradients of $\mathcal{J}$ are clipped to a maximum norm to ensure stable learning dynamics. Hyperparameters $\gamma$ and $\lambda_{\text{GAE}}$ control temporal credit assignment via Generalized Advantage Estimation \ref{eq:GAE}, $\epsilon$ specifies the clipping range for the ratio constraint \ref{eqn:PPO}.

\subsection{Optimal values used across all environments}
\begin{table}[H]
\centering
\begin{tabularx}{\linewidth}{lX}
\hline
\textbf{Parameter} & \textbf{Value} \\
\hline
Coefficient for DClamp-PPO ($\alpha$) & 3 \\
Coefficient for DClamp-PPO ($\beta$) & $\epsilon$ (Optimal clip range for PPO) \\
Coefficient for Leaky-PPO ($\alpha$) & 0.01 \\
Coefficient for PPO-RB ($\alpha$) & 0.02 (Humanoid), 0.3 (Other) \\
\hline
\end{tabularx}
\end{table}

\subsection{Ant-v4}
\begin{table}[H]
\centering
\begin{tabularx}{\linewidth}{lX}
\hline
\textbf{Parameter} & \textbf{Value} \\
\hline
Normalize & True \\
$n_{\text{envs}}$ & 1 \\
Policy & MlpPolicy \\
$n_{\text{timesteps}}$ & $1\times 10^7$ \\
Batch size & 32 \\
$n_{\text{steps}}$ & 512 \\
$\gamma$ & 0.98 \\
Learning rate & $1.90609 \times 10^{-5}$ \\
Entropy coefficient & $4.9646 \times 10^{-7}$ \\
Clip range & 0.1 \\
$n_{\text{epochs}}$ & 10 \\
$\lambda_{\text{GAE}}$ & 0.8 \\
Max grad norm & 0.6 \\
Value function coefficient & 0.677239 \\
\hline
\end{tabularx}
\end{table}

\subsection{HalfCheetah-v4}
\begin{table}[H]
\centering
\begin{tabularx}{\linewidth}{lX}
\hline
\textbf{Parameter} & \textbf{Value} \\
\hline
Normalize & True \\
$n_{\text{envs}}$ & 1 \\
Policy & MlpPolicy \\
$n_{\text{timesteps}}$ & $1\times 10^6$ \\
Batch size & 64 \\
$n_{\text{steps}}$ & 512 \\
$\gamma$ & 0.98 \\
Learning rate & $2.0633 \times 10^{-5}$ \\
Entropy coefficient & $4.01762 \times 10^{-4}$ \\
Clip range & 0.1 \\
$n_{\text{epochs}}$ & 20 \\
$\lambda_{\text{GAE}}$ & 0.92 \\
Max grad norm & 0.8 \\
Value function coefficient & 0.58096 \\
Policy kwargs & log\_std\_init=-2, ortho\_init=False, activation=ReLU, net\_arch=pi=[256,256], vf=[256,256] \\
\hline
\end{tabularx}
\end{table}

\subsection{Hopper-v4}
\begin{table}[H]
\centering
\begin{tabularx}{\linewidth}{lX}
\hline
\textbf{Parameter} & \textbf{Value} \\
\hline
Normalize & True \\
$n_{\text{envs}}$ & 1 \\
Policy & MlpPolicy \\
$n_{\text{timesteps}}$ & $1\times 10^6$ \\
Batch size & 32 \\
$n_{\text{steps}}$ & 512 \\
$\gamma$ & 0.999 \\
Learning rate & $9.80828 \times 10^{-5}$ \\
Entropy coefficient & $2.29519 \times 10^{-3}$ \\
Clip range & 0.2 \\
$n_{\text{epochs}}$ & 5 \\
$\lambda_{\text{GAE}}$ & 0.99 \\
Max grad norm & 0.7 \\
Value function coefficient & 0.835671 \\
Policy kwargs & log\_std\_init=-2, ortho\_init=False, activation=ReLU, net\_arch=pi=[256,256], vf=[256,256] \\
\hline
\end{tabularx}
\end{table}

\subsection{Humanoid-v4}
\begin{table}[H]
\centering
\begin{tabularx}{\linewidth}{lX}
\hline
\textbf{Parameter} & \textbf{Value} \\
\hline
Normalize & True \\
$n_{\text{envs}}$ & 1 \\
Policy & MlpPolicy \\
$n_{\text{timesteps}}$ & $1\times 10^7$ \\
Batch size & 256 \\
$n_{\text{steps}}$ & 512 \\
$\gamma$ & 0.95 \\
Learning rate & $3.56987 \times 10^{-5}$ \\
Entropy coefficient & $2.38306 \times 10^{-3}$ \\
Clip range & 0.3 \\
$n_{\text{epochs}}$ & 5 \\
$\lambda_{\text{GAE}}$ & 0.9 \\
Max grad norm & 2.0 \\
Value function coefficient & 0.431892 \\
Policy kwargs & log\_std\_init=-2, ortho\_init=False, activation=ReLU, net\_arch=pi=[256,256], vf=[256,256] \\
\hline
\end{tabularx}
\end{table}

\subsection{Reacher-v4\protect\footnote{Not provided in RL-Baselines3-Zoo; we used the same configuration as Reacher-v2.}}
\begin{table}[H]
\centering
\begin{tabularx}{\linewidth}{lX}
\hline
\textbf{Parameter} & \textbf{Value} \\
\hline
Normalize & True \\
$n_{\text{envs}}$ & 1 \\
Policy & MlpPolicy \\
$n_{\text{timesteps}}$ & $1\times 10^6$ \\
Batch size & 32 \\
$n_{\text{steps}}$ & 512 \\
$\gamma$ & 0.9 \\
Learning rate & $1.04019 \times 10^{-4}$ \\
Entropy coefficient & $7.52585 \times 10^{-8}$ \\
Clip range & 0.3 \\
$n_{\text{epochs}}$ & 5 \\
$\lambda_{\text{GAE}}$ & 1.0 \\
Max grad norm & 0.9 \\
Value function coefficient & 0.950368 \\
\hline
\end{tabularx}
\end{table}

\subsection{Walker2d-v4}
\begin{table}[H]
\centering
\begin{tabularx}{\linewidth}{lX}
\hline
\textbf{Parameter} & \textbf{Value} \\
\hline
Normalize & True \\
$n_{\text{envs}}$ & 1 \\
Policy & MlpPolicy \\
$n_{\text{timesteps}}$ & $1\times 10^6$ \\
Batch size & 32 \\
$n_{\text{steps}}$ & 512 \\
$\gamma$ & 0.99 \\
Learning rate & $5.05041 \times 10^{-5}$ \\
Entropy coefficient & $5.85045 \times 10^{-4}$ \\
Clip range & 0.1 \\
$n_{\text{epochs}}$ & 20 \\
$\lambda_{\text{GAE}}$ & 0.95 \\
Max grad norm & 1.0 \\
Value function coefficient & 0.871923 \\
\hline
\end{tabularx}
\end{table}

\subsection{Swimmer-v4}
\begin{table}[H]
\centering
\begin{tabularx}{\linewidth}{lX}
\hline
\textbf{Parameter} & \textbf{Value} \\
\hline
Policy & MlpPolicy \\
$n_{\text{envs}}$ & 4 \\
$n_{\text{timesteps}}$ & $1\times 10^6$ \\
$n_{\text{steps}}$ & 1024 \\
Batch size & 256 \\
$n_{\text{epochs}}$ & 10 \\
$\gamma$ & 0.9999 \\
$\lambda_{\text{GAE}}$ & 0.98 \\
Clip range & 0.2 \\
Entropy coefficient & 0.0 \\
Value function coefficient & 0.5 \\
Max grad norm & 0.5 \\
Learning rate & $6\times 10^{-4}$ \\
\hline
\end{tabularx}
\end{table}

\newpage
\section{Additional Experimental results}
\label{App:addExp}
The following figures present additional training results following the same evaluation protocol described in the main text (see Figure~\ref{fig:all_envs}).
Each curve reports the mean episodic return over 10 evaluation episodes, averaged across 5 random seeds, with shaded regions showing one standard deviation.
The results are smoothed using a moving-average window of 10.
These supplementary plots include additional environments and ablations over the $\alpha$ and $\beta$ coefficients used in the DClamp-PPO loss formulation.
\begin{figure}[H]
\centering

\begin{minipage}[t]{0.45\linewidth}
  \centering
  \includegraphics[width=\linewidth,height=5cm,keepaspectratio]{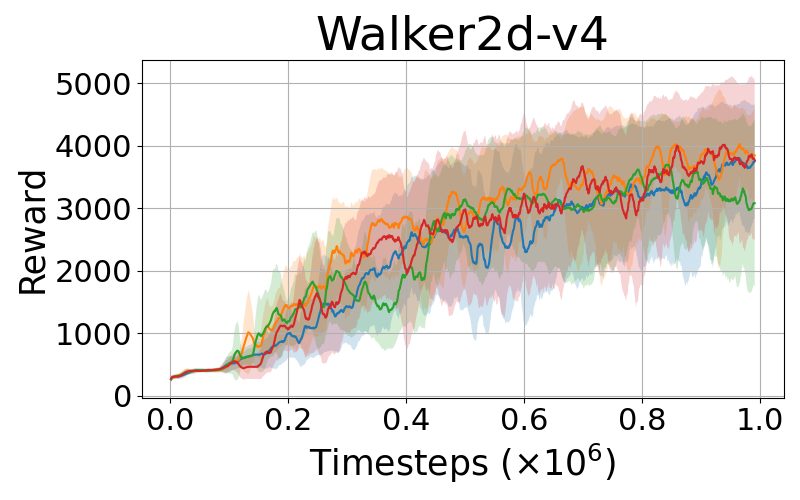}
\end{minipage}\hfill
\begin{minipage}[t]{0.45\linewidth}
  \centering
  \includegraphics[width=\linewidth,height=5cm,keepaspectratio]{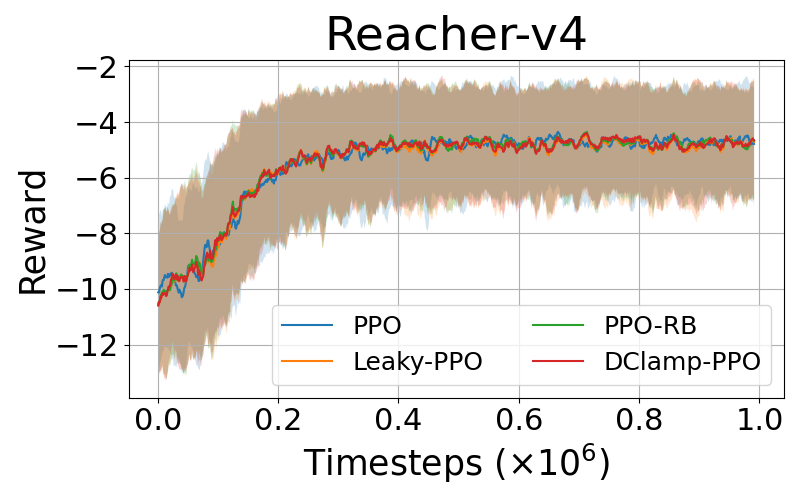}
\end{minipage}

\vspace{0.5cm} 

\begin{minipage}[t]{0.45\linewidth}
  \centering
  \includegraphics[width=\linewidth,height=5cm,keepaspectratio]{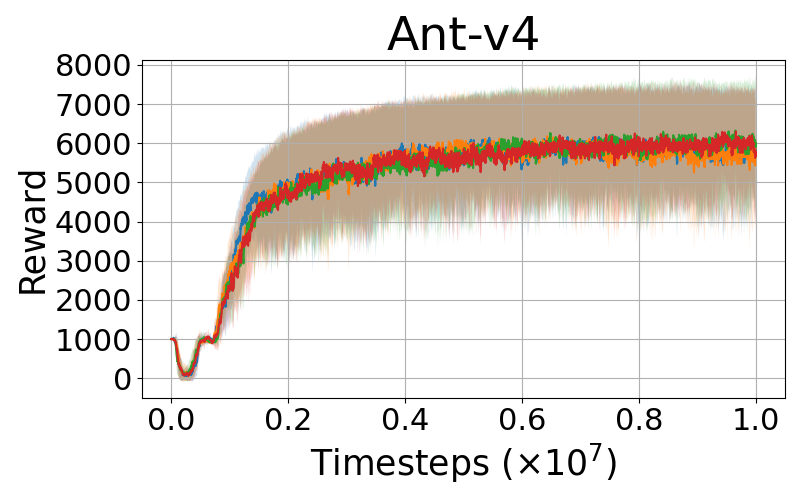}
\end{minipage}

\caption{Performance comparisons of PPO, PPO-Leaky, PPO-RB and DClamp-PPO (ours) across MuJoCo environments.}
\label{fig:add}
\end{figure}
By Figure \ref{fig:add}, across all three environments, DClamp-PPO (red) consistently performs on par with or slightly better than other variants, while also showing more stable learning dynamics. In Walker2d-v4, both DClamp-PPO and Leaky-PPO achieve faster initial improvement and higher final rewards compared to other variants. In Reacher-v4, performance differences are smaller, but DClamp-PPO maintains robustness. In Ant-v4, all methods reach similar asymptotic performance, though DClamp-PPO and PPO-RB display smoother convergence compared to PPO. 
\begin{figure}[H]
\centering
\begin{minipage}[t]{0.48\linewidth}
  \centering
  \includegraphics[width=\linewidth, keepaspectratio]{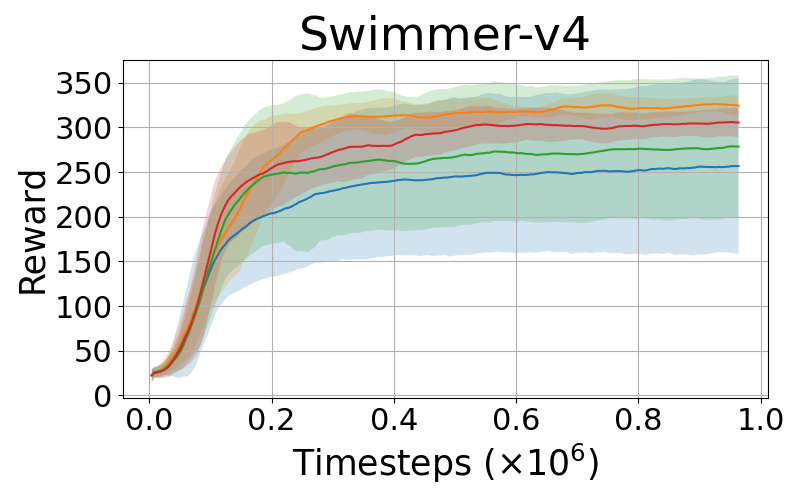}
\end{minipage}
\hfill
\begin{minipage}[t]{0.48\linewidth}
  \centering
  \includegraphics[width=\linewidth, keepaspectratio]{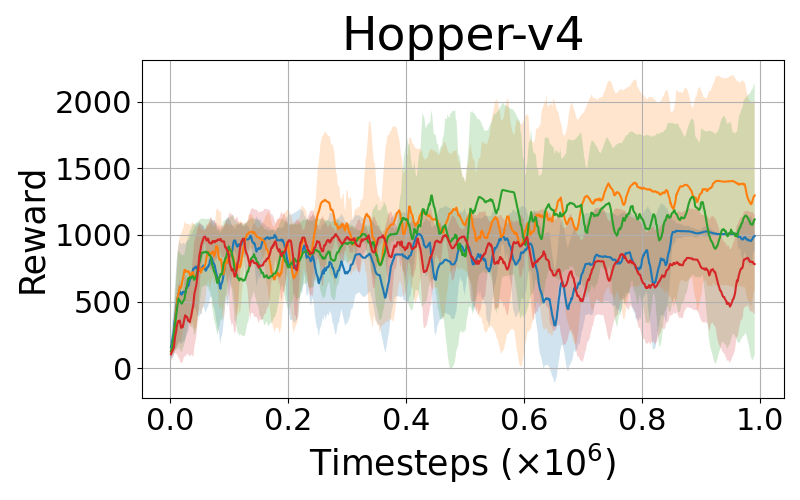}
\end{minipage}

\vspace{0.5em} 

\begin{minipage}[t]{0.48\linewidth}
  \centering
  \includegraphics[width=\linewidth, keepaspectratio]{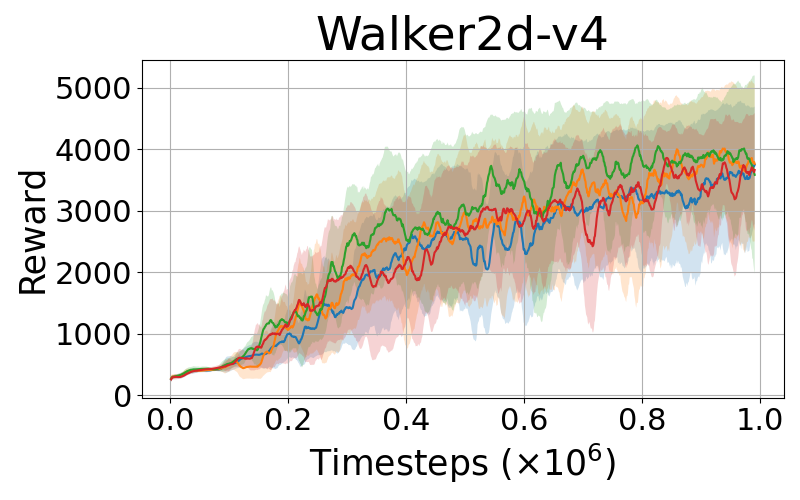}
\end{minipage}
\hfill
\begin{minipage}[t]{0.48\linewidth}
  \centering
  \includegraphics[width=\linewidth, keepaspectratio]{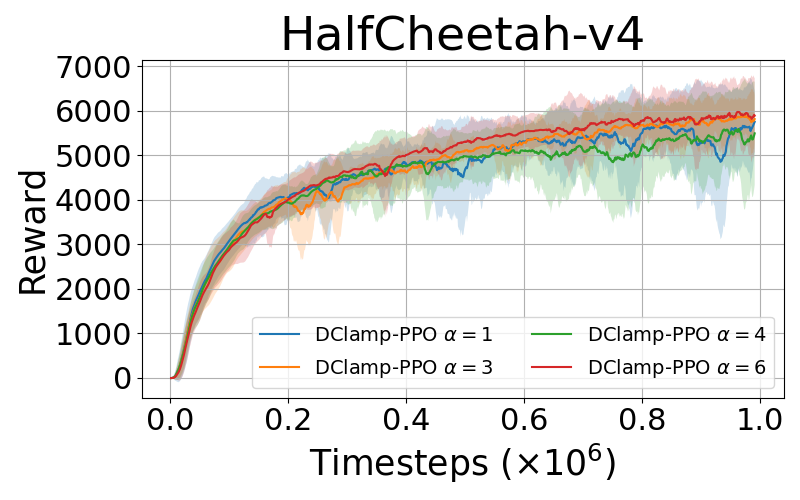}

\end{minipage}

\caption{Performance comparisons with different coefficients of $\beta$ across MuJoCo environments using the optimal coefficient $\beta$ provided in~\cref{app:hyperparams} averaged over 5 random seeds.}

\label{fig:diffalpha}
\end{figure}
Figure~\ref{fig:diffalpha} illustrates the sensitivity of DClamp-PPO to the coefficient $\alpha$ across several MuJoCo environments. 
Overall, the performance remains relatively stable for moderate values of $\alpha$, demonstrating the robustness of the method. 
In Swimmer-v4 and HalfCheetah-v4, higher $\alpha$ values mostly improve convergence speed and final rewards, 
indicating that stronger directional clamping can facilitate more decisive updates. 
Conversely, in Hopper-v4 and Walker2d-v4, overly large $\alpha$ occasionally leads to higher variance and slower stabilization, 
suggesting that moderate clamping (e.g., $\alpha=3$–$4$) provides a better trade-off between exploration and stability. 
Overall, DClamp-PPO exhibits strong resilience to hyperparameter tuning.

\begin{figure}[H]
\centering
\begin{minipage}[t]{0.48\linewidth}
  \centering
  \includegraphics[width=\linewidth, keepaspectratio]{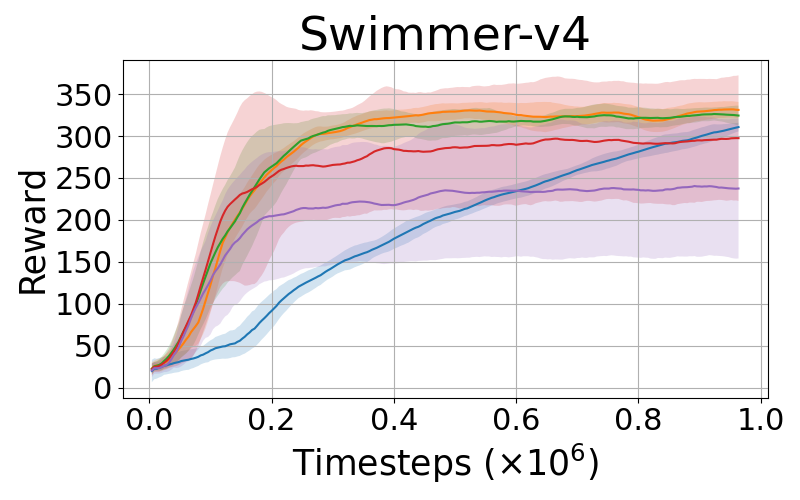}
\end{minipage}
\hfill
\begin{minipage}[t]{0.48\linewidth}
  \centering
  \includegraphics[width=\linewidth, keepaspectratio]{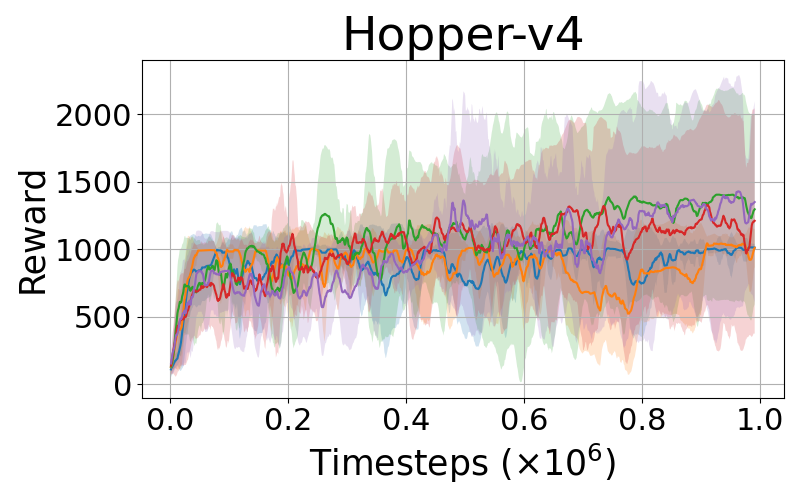}
\end{minipage}

\vspace{0.5em} 
\begin{minipage}[t]{0.48\linewidth}
  \centering
  \includegraphics[width=\linewidth, keepaspectratio]{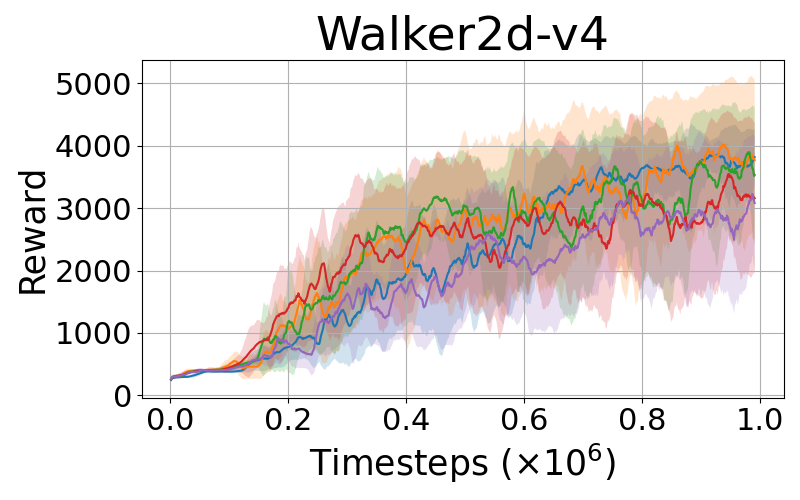}
\end{minipage}
\hfill
\begin{minipage}[t]{0.48\linewidth}
  \centering
  \includegraphics[width=\linewidth, keepaspectratio]{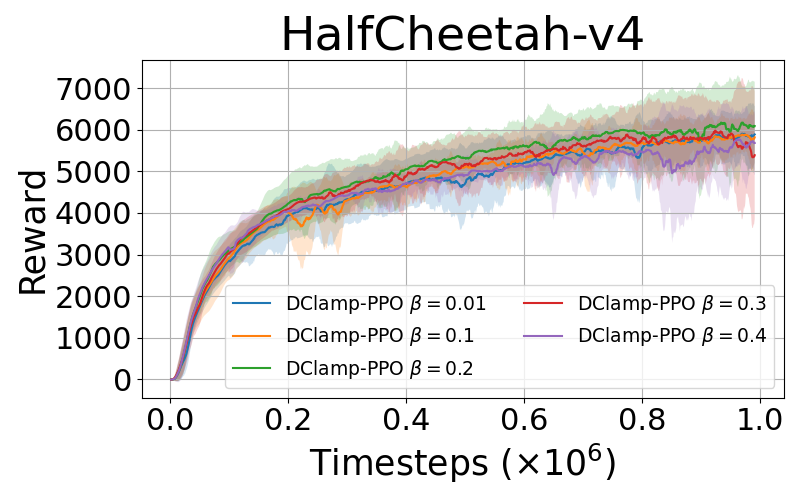}
\end{minipage}

\caption{Performance with different $\beta$ coefficients across MuJoCo environments using the optimal $\alpha$ from~\cref{app:hyperparams} averaged over 5 random seeds.}
\label{fig:diffbeta}
\end{figure}

Figure~\ref{fig:diffbeta} illustrates the effect of varying the coefficient $\beta$ in DClamp-PPO across multiple MuJoCo environments.
Overall, $\beta$ primarily controls the strength of directional correction.
Smaller values (e.g., $\beta=0.01$) impose stronger regularization, which can occasionally slow convergence or introduce mild instability, as observed in Swimmer-v4.
In contrast, larger $\beta$ values yield weaker directional constraints, effectively causing the method to behave increasingly like standard PPO, where ratio deviations are no longer further penalized.
Across environments such as HalfCheetah-v4 and Swimmer-v4, intermediate values ($\beta \in [0.1, 0.2]$) generally provide the best balance between stability and learning speed.
These results confirm that DClamp-PPO remains robust to a wide range of~$\beta$ settings, maintaining stable and competitive performance throughout.
\newpage
In addition to the training curves, Table~\ref{tab:wrong-band-comparison} presents a quantitative comparison of the proportion of ratio samples that fall within the wrong-direction bands for both positive and negative advantages.
The proportions are computed over all state–action transitions $(s,a)$ collected during training. 

\begin{table}[H]
\centering

\begin{tabular}{lcccccc}
\toprule
\textbf{Environment} 
& \multicolumn{2}{c}{DClamp-PPO ($\beta=0.01$)} 
& \multicolumn{2}{c}{DClamp-PPO (Optimal $\beta$)} 
& \multicolumn{2}{c}{PPO} \\ 
\cmidrule(lr){2-3}\cmidrule(lr){4-5}\cmidrule(lr){6-7}
& $\hat{A}(s_t,a_t) < 0$ & $\hat{A}(s_t,a_t) > 0$ 
& $\hat{A}(s_t,a_t) < 0$ & $\hat{A}(s_t,a_t) > 0$ 
& $\hat{A}(s_t,a_t) < 0$ & $\hat{A}(s_t,a_t) > 0$ \\
\midrule
HalfCheetah-v4 & \textbf{0.2701} & \textbf{0.3109} & 0.3126 & 0.3636 & 0.3092 & 0.3755 \\
Hopper-v4      & 0.4484 & \textbf{0.4115} & 0.4017 & 0.4638 & \textbf{0.3923} & 0.4776 \\
Swimmer-v4     & 0.4927 & \textbf{0.4551} & 0.4895 & 0.4772 & \textbf{0.4682} & 0.4931 \\
Walker2d-v4    & 0.3487 & 0.3646 & 0.3474 & 0.3749 & \textbf{0.3303} & \textbf{0.3610} \\
Ant-v4         & \textbf{0.3432} & \textbf{0.3668} & 0.3568 & 0.3901 & 0.3500 & 0.3866 \\
Humanoid-v4    & \textbf{0.3096} & \textbf{0.3595} & 0.3402 & 0.4129 & 0.3355 & 0.4139 \\
Reacher-v4     & 0.4644 & \textbf{0.4578} & 0.4369 & 0.4879 & \textbf{0.4244} & 0.4928 \\
\midrule
\textbf{Overall mean} 
& 0.3824 & \textbf{0.3895} 
& 0.3836 & 0.4243 
& \textbf{0.3728} & 0.4286 \\
\bottomrule
\end{tabular}
\caption{Comparison between DClamp-PPO and PPO in terms of the proportion of samples in the wrong-direction band for positive ($\hat{A}(s_t,a_t)>0$) and negative ($\hat{A}(s_t,a_t)<0$) advantages across MuJoCo environments (with the optimal $\alpha, \beta$ given in~\cref{app:hyperparams}). Bold indicates the lowest proportion of samples in wrong-direction band out of the three.}
\label{tab:wrong-band-comparison}
\end{table}

Table~\ref{tab:wrong-band-comparison} compares the proportion of ratio samples that fall within the wrong-direction bands for both positive and negative advantages. 
Across most MuJoCo environments, DClamp-PPO (with $\beta=0.01$) consistently reduces the frequency of wrong-direction updates compared to standard PPO. 
The overall mean confirms this trend—DClamp-PPO reduces wrong-direction updates by roughly 4\% on average relative to PPO, 
demonstrating that directional clamping could effectively mitigate wrong-direction updates.
\end{document}